\documentclass{article}
\usepackage{ijcai25}
\usepackage{amsmath,amsfonts}
\usepackage{subfigure}
\usepackage{times}
\usepackage{soul}
\usepackage{url}
\usepackage[hidelinks]{hyperref}
\usepackage[utf8]{inputenc}
\usepackage[small]{caption}
\usepackage{graphicx}
\usepackage{amsmath}
\usepackage{amsthm}
\usepackage{booktabs}
\usepackage{algorithm}
\usepackage{algorithmic}
\usepackage[switch]{lineno}
\usepackage{multirow}
\usepackage{bbding}
\usepackage{makecell}

\title{Few-shot Novel Category Discovery}

\author{
Chunming Li$^1$
\and
Shidong Wang$^2$\and
Haofeng Zhang$^1$\thanks{Corresponding Author.}
\affiliations
$^1$School of Computer Science and Engineering, Nanjing University of Science and Technology, China.\\
$^2$School of Engineering, Newcastle University, Newcastle upon Tyne, United Kingdom.
\emails
\{chunmingli, zhanghf\}@njust.edu.cn,
shidong.wang@newcastle.ac.uk
}

\begin{document}

\maketitle

\begin{abstract}
The recently proposed Novel Category Discovery (NCD) adapt paradigm of transductive learning hinders its application in more real-world scenarios. In fact, few labeled data in part of new categories can well alleviate this burden, which coincides with the ease that people can label few of new category data. Therefore, this paper presents a new setting in which a trained agent is able to flexibly switch between the tasks of identifying examples of known (labelled) classes and clustering novel (completely unlabeled) classes as the number of query examples increases by leveraging knowledge learned from only a few (handful) support examples. Drawing inspiration from the discovery of novel categories using prior-based clustering algorithms, we introduce a novel framework that further relaxes its assumptions to the real-world open set level by unifying the concept of model adaptability in few-shot learning. We refer to this setting as Few-Shot Novel Category Discovery (FSNCD) and propose Semi-supervised Hierarchical Clustering (SHC) and Uncertainty-aware K-means Clustering (UKC) to examine the model's reasoning capabilities.
Extensive experiments and detailed analysis on five commonly used datasets demonstrate that our methods can achieve leading performance levels across different task settings and scenarios. Code is available at: \url{https://github.com/Ashengl/FSNCD}.
\end{abstract}

\section{Introduction}
Most deep learning methods \cite{resnet,MAE,ViT} aim to excel in supervised learning, where models assign labels to test samples using knowledge from training. However, this approach poorly reflects real-world scenarios, where unlabelled data may include both known and novel classes. Few-shot Learning (FSL) \cite{finn2017model,sung2018learning,fei2006one,vinyals2016matching,li2023libfewshot} addresses this by mimicking human ability to learn from limited examples. Despite its potential, FSL has a critical drawback: if not all categories are annotated, it forces classification of unseen classes into existing ones, contradicting open-world reasoning.

The advent of Novel Category Discovery (NCD) \cite{openldn} and Generalized Category Discovery (GCD) \cite{GCD} lifts the learning agent to a new level, where the novel examples are allowed to be grouped into a few clusters based on their inherent characteristics. Notwithstanding, the key of these two tasks will degenerate into solving a cross-domain clustering problem if unlabelled data are not presented. A recent research, On-the-fly Category Discovery (OCD) \cite{du2023on} was proposed to mitigate the limitation of transductive learning involved in NCD and GCD by determining novel classes according to the values derived from decoupling features into symbolic features and absolute features. However, the over-relaxed constraints imposed on the hash encoding for novel classes lead to a phenomenon in which a higher number of clusters formed than initial expectation.

 \begin{figure*}[t]
	\centering
	\includegraphics[width=0.99\textwidth]{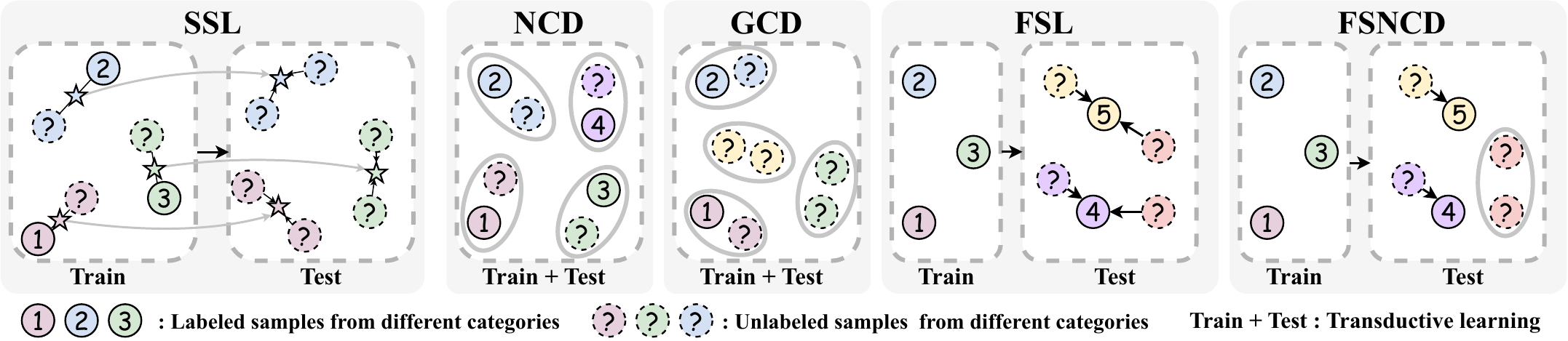}
  \vspace{-2ex}
	\caption{Comparison of Different Task Settings. Solid lines represent labeled data, dashed lines represent unlabeled data, with different colors indicating different categories. Unlabeled data in test phase are samples to be classified. Inductive learning and transductive learning are annotated as ``Train/Test'' and ``Train+Test''. Please note that during the generalization testing phase in inductive learning, FSL and FSNCD are based on a small amount of labeled data.} 
	\label{fig_pre}
 \vspace{-2ex}
\end{figure*}

To enhance the adaptability of learning agent to real-world scenarios, we adapt the assumptions and training strategies in NCD. Similar to FSL, we ensure no overlap between training and test label spaces, and the test set's query label space extends beyond the support set to include novel categories. Unlike semi-supervised strategies in NCD \cite{openldn} and GCD \cite{GCD}, we adopt episodic meta-learning to better evaluate the model's adaptability to new tasks.

We call this new problem Few-Shot Novel Category Discovery (FSNCD), and it is intersected conception at the FSL \cite{fei2006one} and NCD \cite{openldn}, empowering a learning agent with the capability to be inductively adaptive to the discovery of both novel classes and clusters. The key setting of this problem is illustrated in Fig. \ref{fig_pre}, and it builds upon the existing setting of NCD \cite{openldn} which was introduced to cluster novel classes by hints of both labelled and unlabelled data. This scenario assumes that a learning agent can both extract features from labeled data and learn patterns from unlabeled data. Our proposed setting challenges this by restricting access to unlabeled data, focusing instead on enhancing the model's adaptability and plasticity to improve generalization.

FSNCD leverages a DINO \cite{dino} pre-trained ViT-B \cite{ViT} model, differing from traditional few-shot learning by tackling novel category discovery, which involves clustering novel classes rather than simple detection. We introduce a hierarchical clustering method with a stopping criterion based on inter-class distances of visible classes, addressing bias from excessive novel categories through balanced sparsity. Additionally, an improved K-means algorithm incorporates uncertainty clustering to handle distributions. Unlike existing methods reliant on prior category counts, our approach restructures K-means and hierarchical clustering to flexibly estimate category numbers in few-shot settings, facilitating the discovery of new categories.

To sum up, our contributions are as follows:
\begin{itemize}
    \item We propose a new task setting named Few-Shot Novel Category Discovery (FSNCD). It unifies few-shot learning and Novel Category Discovery, realising a selectivity between tasks of identifying examples from labelled classes in a real-time inference manner and clustering novel classes as the query examples increase.   
    \item We introduce two clustering algorithms as baseline methods: Semi-supervised Hierarchical Clustering (SHC) and Uncertainty-aware K-means Clustering (UKC). Both approaches take into account the specific features required to address the complexities of open-set recognition, extending beyond conventional data clustering techniques.
    \item We conducted experiments on five datasets, analyzing and elucidating the viability of the proposed new setting.
\end{itemize}

\section{Related work}
\label{sec_rw}
\subsection{Semi-supervised learning}
In traditional supervised learning, we need a large amount of labeled data to train the model \cite{SSL0,SSLsurvey}. To address the problem of insufficient labeled samples, semi-supervised learning is proposed to utilize labeled and unlabeled data to improve model performance with limited labeled data \cite{NIPS2004_96f2b50b,4664785,semi_tnn1,semi_tnn2}. The most intuitive approach is perhaps Pseudo-Labels self-training \cite{PL}, where a model trained on labeled data generates categorical pseudo-labels for the unlabeled examples. SSL techniques often utilize unlabeled data in various ways, such as employing strategies like consistency regularization \cite{meanteacher}, generative model based methods \cite{8363749}, or graph based method \cite{10113741}.

\subsection{Novel category discovery}
Novel Category Discovery (NCD) has recently been proposed to address the challenge of limited labeled data by automatically classifying unlabeled data using labelled samples. Initially normalized by DTC \cite{openldn}, NCD has gained widespread attention. AutoNovel \cite{autonovel1} proposes using self-supervised pretraining and generates pseudo-labels through ranking, training the model alongside the ground truth of labeled data.

Recently, tasks related to NCD have garnered increasing attention from researchers. Novel Category Discovery in semantic segmentation \cite{ncdss} aims to differentiate objects and backgrounds while discovering novel categories. Novel Category Discovery based on incremental learning \cite{restune,incd} proposes the discovery of novel categories while maintaining the ability to recognize previously seen ones. With further research, Vaze et al. introduces GCD \cite{GCD}, which requires models trained on a few labeled categories to recognize known and novel categories from newly collected unlabeled data. SimGCD \cite{wen2023simgcd} introduces parametric classification tailored for the GCD task. To address the real-time inference for detecting new classes, OCD \cite{du2023on} proposes an efficient hash coding scheme.

\subsection{Few-shot learning}
Few-shot learning (FSL) is a method used to address the issue of limited sample size \cite{fei2006one,finn2017model,snell2017prototypical,fsl1,9678031}. It aims to learn an effective classification model from a few labeled training examples. There are two main types of FSL methods: meta-learning-based  \cite{finn2017model,ravi2016optimization,9921329} and metric-learning-based approaches \cite{snell2017prototypical,koch2015siamese}. Additionally, some methods employ a Non-episodic-based strategy \cite{chen2019closerfewshot,Dhillon2020A}, i.e., fine-tuning a pre-trained model at test time. Although current methods for Few-shot Open-set Recognition \cite{fsosr1,fsosr2} are capable of detecting out-of-distribution samples, they lack the ability to cluster or classify these out-of-distribution samples. 

\section{Few-shot Novel Category Discovery}
\label{sec_med}

\begin{table}[!t]
\vspace{-2ex}
\centering
\resizebox{0.49\textwidth}{!}{
\begin{tabular}{lccccc} \\ \toprule
Method & Training & Test & Transferring Manner & Support & Query \\\midrule
SSL & $\mathcal{Y}^\text{L}\cup\mathcal{Y}^\text{L}$$^\dag$ & $\mathcal{Y}^\text{L}$ & Inductive & - & - \\
NCD & $\mathcal{Y}^\text{L}\cup\mathcal{Y}^\text{U}$$^\dag$ & - & Transductive & - & - \\
GCD & $\mathcal{Y}^\text{L}\cup\mathcal{Y}^\text{U}$$^\dag\cup\mathcal{Y}^\text{L}$$^\dag$ & - & Transductive & - & - \\
FSL & $\mathcal{Y}^\text{L}$ & $\mathcal{Y}^\text{U}$ & Inductive &   $\mathcal{Y}^\text{S}$ & $\mathcal{Y}^\text{S}$\\
\textbf{FSNCD} & $\mathcal{Y}^\text{L}$ & $\mathcal{Y}^\text{U}$ & Inductive & $\mathcal{Y}^\text{S}$ & $\mathcal{Y}^\text{S}\cup\mathcal{Y}^\text{N}$ \\ \bottomrule
\end{tabular}}
\caption{Comparison between different tasks. \dag denotes providing images during the training phase without accompanying labels.}
\label{tab_tasks}
\end{table}

\subsection{Problem definition}
\noindent\textbf{NCD:} The training data for NCD is provided in two distinct sets, including a labelled set $\mathcal{D}^\text{L}=\{(x_i^l,y_i^l)\}_{i=1}^N$ and an unlabeled set $\mathcal{D}^\text{U}=\{x_i^u\}_{i=1}^M$, where an instance is denoted as $x_i^l$ or $x_i^u$ depending on which set the data is derived from, $y_i^l \in \mathcal{Y}^\text{L}=\{1, \dots, C^\text{L}\}$ is the corresponding class label of $\mathcal{D}^\text{L}$ and $\mathcal{Y}^\text{U}$ is the label space of $\mathcal{D}^\text{U}$. The goal is to train a model on both $\mathcal{D}^\text{L}$ and $\mathcal{D}^\text{U}$, and then aggregate $\mathcal{D}^\text{U}$ into $C^\text{U}$ clusters that are associated with $\mathcal{Y}^\text{U}=\{1, \dots, C^\text{U}\}$ where $\mathcal{Y}^\text{L}\cap \mathcal{Y}^\text{U}=\emptyset$. GCD further relaxes the assumptions of NCD, with an emphasis on identifying both labelled data and novel categories, where the label space of $\mathcal{D}^\text{U}$ changes to $\mathcal{Y}^\text{L}\cup\mathcal{Y}^\text{U}$ instead of $\mathcal{Y}^\text{U}$.


\noindent\textbf{FSL:} In FSL, a dataset $D$ is usually partitioned into two sets, the training set containing base classes $\mathcal{D}^{\text{Base}}$ and the test set containing novel classes $\mathcal{D}^{\text{Novel}}$. It creates $N$ training tasks $\mathcal{T}=\{T_1, \dots, T_N\}$ to assemble a finite set of training episodes where each training task $\mathcal{T}_{n}$ consists of different classes, represented by $\mathcal{T}_{n} = <\mathcal{S}, \mathcal{Q}>$, where $\mathcal{S}$ and $\mathcal{Q}$ denote examples chosen from the support set and query set, respectively. During testing, the model receives new tasks with the novel classes which have no overlap with those encountered during training.     

\noindent\textbf{FSNCD:} We approach Few-Shot Novel Category Discovery as a scenario where the dataset is organized under the setting defined in FSL with training set containing base classes $\mathcal{D}^{\text{Base}}$ and the test set containing novel classes $\mathcal{D}^{\text{Novel}}$. However, an episode in FSNCD contains a query set $\mathcal{Q} = \{(x^q, y^q)\} \in \mathcal{X} \times \mathcal{Y}^\text{Q}$, and a support set $\mathcal{S} = \{(x_l^{s}, y_l^{s})\}^{NK}_{l=1} \in \mathcal{X} \times \mathcal{Y}^\text{S}$ of $K$ image-label pairs for each $N$ classes, referring to an $N$-way $K$-shot setting. However, the label space in FSNCD is defined as $\mathcal{Y}^{\text{Q}}=\mathcal{Y}^\text{S}\cup\mathcal{Y}^\text{N}$ rather than $\mathcal{Y}^{\text{Q}} = \mathcal{Y}^{\text{S}}$ as defined in standard FSL. The primary objective for the learning agent therefore is to identify examples of categories in $\mathcal{Y}^{\text{S}}$ while discovering unknown categories in $\mathcal{Y}^\text{N}$. 

The characteristics of the proposed setting are summarised in Table \ref{tab_tasks} and Fig. \ref{fig_pre}. SSL is an inductive approach leveraging both labeled and unlabeled data within the same label space for test set classification. GCD relaxes the closed-set assumption, allowing the label space of unlabeled data to encompass that of labeled data, following a transductive paradigm. Transitioning from NCD and GCD to FSL introduces FSNCD, which combines few-shot classification with the discovery of novel categories. A key distinction between FSNCD and traditional FSL is that in the testing phase of FSNCD, certain samples may not match any of the support categories. Unlike Few-Shot Open Set Recognition (FSOSR), which only detects outliers, FSNCD requires models to detect and accurately cluster these outliers. Additionally, in contrast to open-world recognition, categories appearing in testing phase of FSNCD are not seen during training, though a limited number of samples are provided during testing.

\subsection{Training of FSNCD}
\begin{figure*}[!t]
\centering
\includegraphics[width=0.96\textwidth]{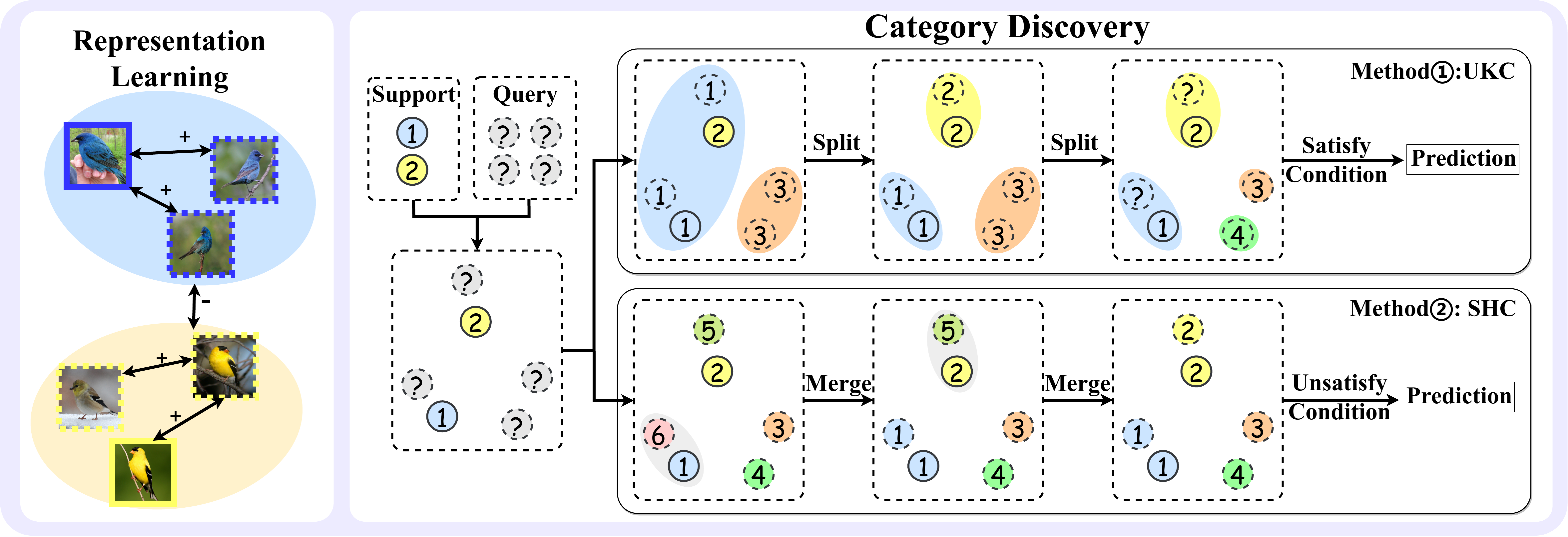} 
 \vspace{-2ex}
\caption{Illustration of our proposed baselines. In the representation learning stage, each episode takes both support and query samples into a min-batch selected from  $\mathcal{D}^\text{L}$ and trains a feature extractor by using the supervised contrastive learning approach. In the second stage, we test on $\mathcal{D}^{U}$, and for each episode, we utilize the proposed clustering method to classify samples belonging to $\mathcal{Y}^\text{S}$ and cluster new categories.} 
\label{fig_intro}
\vspace{-2ex}
\end{figure*}

The training procedure for a learning agent in FSNCD is divided into two consecutive phases: representation learning and classifier construction. For the former phase, the learning model is dedicated to training a feature extractor $\mathcal{F}$ and a projection head $\phi$, $\phi \circ \mathcal{F}: \mathcal{X} \rightarrow \mathcal{Z}$, with $\mathcal{Z}$ signifying the embedding space generated from input images $\mathcal{X}$ via $\phi \circ \mathcal{F}$. The latter phase is responsible for constructing a classifier capable of recognizing query examples and discovering novel categories. An overview of our approach is demonstrated in Fig. \ref{fig_intro}. A detailed explanation of each phase will be presented in the sequel below. 

\noindent\textbf{Phase 1: Representation learning}\label{phs1}

\noindent Given that the training process only involves the use of labelled data, the supervised contrastive learning loss \cite{khosla2020supervised} is employed to learn more distinctive features for samples belonging to the same category and different categories within each batch. This mitigates the impact of using standard cross-entropy loss for prototype matching which is classification-prone rather than concentrating on the more critical clustering task, and avoids the occurrence of the visible-class bias described in ProtoNet \cite{snell2017prototypical}. Specifically, given an input image $\mathbf{x}$, it uses the feature extractor $\mathcal{F}\left(\mathbf{x}\right)$, together with a multilayer perceptron (MLP) projection $\phi$ to extract the feature embeddings $\mathbf{z}$, denoted as $\mathbf{z}=\phi\left(\mathcal{F}\left(\mathbf{x}\right)\right)$. Formally,  
\begin{equation}
    \mathcal{L}_{SL}=\sum_{i\in E}\left(-\frac{1}{\left |\mathcal{N}(i)\right |}\sum_{q\in \mathcal{N}(i)}\operatorname{log}\frac{\operatorname{exp}(\mathbf{z}_i\cdot \mathbf{z}_q/\tau)}{\sum_{n}\mathbb{I}_{[n\neq i]}\operatorname{exp}(\mathbf{z}_i\cdot \mathbf{z}_n /\tau)}\right),
\end{equation}
where $\mathcal{N}(i)$ signifies the images belonging to the same category as the input image $\mathbf{x}_i$ in an episode, where $\mathbb{I}$ is an indicator function evaluating to 1 \textit{iff} $ n\neq i$. 

\noindent\textbf{Phase 2: Classifier construction}

\noindent Using the previously learned representations, the next task is to construct classifiers with the ability to classify query samples into known classes or group a set of examples into new categories. Due to our aim to propose solutions addressing more realistic problems, the varying number of novel categories and the uncertain number of samples per novel category make it hard to construct a fixed-parameterized classifier. Thus, two clustering algorithms, Semi-supervised Hierarchical Clustering (SHC) and Uncertainty-aware K-means Clustering (UKC), are specifically designed for FSNCD scenarios, enhancing data grouping in open-set scenarios by aligning more closely with their inherent properties.

\noindent\textbf{Semi-supervised Hierarchical Clustering.}
Hierarchical clustering is one of the most commonly used approaches for grouping data, distinguished by its ability to determine the number of clusters with no need to pre-specify the number of clusters. Grouped clusters that are visually represented in a hierarchical tree called a \textit{dendrogram} whose structural characteristics are the key to achieving optimal clustering. However, it is always a challenge in practice to determine the best criteria to end the accumulation or division during clustering.    

To tackle this issue, we tend to make the most of the information represented by features extracted from the support set which are capable of explicitly characterizing discrepancies between multiple class prototypes. The idea behind this is that hierarchical clustering should cease as chosen prototypes that originate from different classes are merged. Concretely, we introduce an agglomerative hierarchical clustering approach, inspired by the Unweighted Pair Group Method with Arithmetic Mean (UPGMA) where the closest clusters are combined into a higher-level cluster at each step. For each step, the distance between any two clusters $C_i$ and $C_j$ is calculated by averaging the distance between elements within the cluster, expressed by,      
\begin{equation}
    d_{C_i, C_j} = \frac{1}{\left|C_i\right|\cdot\left|C_j\right|}\sum_{x\in C_i}\sum_{y\in C_j}d(x,y),
\end{equation}
where $d(*,*)$ denotes the distance function. In each iteration, the two clusters with the shortest distance are merged, and the clustering between all clusters is updated by,
\begin{equation}
    d_{(C_i\cup C_j), C_k} = \frac{\left|C_i\right|\cdot d_{C_i, C_k}+\left|C_j\right|\cdot d_{C_j, C_k}}{\left|C_i\right| + \left|C_j\right|}.
\end{equation}
The proposed clustering aims at merging the most similar classes according to the adopted cosine metric and triggers iteration termination until two different class prototypes are grouped. Furthermore, clustering is conditioned on the number of query samples, i.e., it will be considered a potential cluster $P$ only if the amount of included samples exceeds a certain value, in which case the remaining clusters are designated as $R$. Moreover, in case there are support prototypes that are not assigned to any cluster, they will be incorporated into the potential class $P$ at the end. With the available class prototypes of the potential classes, the remaining samples will be classified, demonstrating a flexible method of classifying or clustering query samples, called semi-supervised hierarchical clustering.

\noindent\textbf{Uncertainty-aware K-means Clustering.}
K-means clustering is an effective method to iteratively partition a certain number of data observations into $K$ clusters, but the need for the pre-specified number of clusters always hinders its generalization to scenarios full of uncertainty. The adoption of K-means on synthetic experimental datasets is based on a \textit{prior} on how human perception of the $K$ values, which is unrealistic when dealing with real-world tasks where such values are difficult to determine. In addition to the heavy reliance on prior knowledge, the performance of standard K-means clustering strongly depends on the choice of the initialized cluster centroids, which is another uncertainty that needs to be addressed. K-means++ \cite{arthur2007kmeans} presents a stepwise selection of cluster centers, primarily motivated by the idea that samples farther away from other cluster centers are more likely to be selected as the next cluster center. However, the initial number of categories is still an indispensable prior information.

We introduce a new strategy that builds on the assumption that when clustering is performed in the proposed FSNCD setting, both support and query examples lie in a uniform, high-dimensional feature space, and the uncertainty estimates of this space facilitate the choice of the number of centroids implied by novel categories. Specifically, we consider that the features of prototypes and query examples are located in a shared space, denoted by $\mathbf{z}$, and represent standard K-means as $K{\text -}means \left(\cdot, \cdot, \cdot\right)$, then the corresponding potential cluster $P$ can be achieved by,
\begin{equation}
    P = K{\text -}means\left(\mathbf{z}, n, c\right),
\end{equation}
where $n$ is number of categories initialized with $\left|\mathcal{Y_S}\right|$, $c$ denotes the centroids and is initialized randomly. As the iteration $k$ increases, the average quantity is calculated based on the previous clustering results, denoted as $m=\frac{1}{\left|P^k\right|}\sum_{i=1}^{\left|P^k\right|}{\left|P_i^k\right|}$. The estimation of the number of categories is then transformed into an acceptance criterion for a particular class in $P_i^k$ that surpasses the average quantity $m$, denoted as $\alpha$. We then need to split all potential classes where each class is assigned a category count by,
\begin{equation}
n_{P_i^k}=\left\{
\begin{array}{rcl}
\delta(P_i^k) & \hspace{0.3cm} & {\delta(P_i^k) \ge 2}\\
2     & \hspace{0.3cm} & {\delta(P_i^k) < 2 \And \mu(P_i^k)\ge\alpha m}\\
1     & \hspace{0.3cm} & {\delta(P_i^k) < 2 \And \mu(P_i^k)<\alpha m}
\end{array} \right. ,
\end{equation}
where $\delta(\cdot)$ denotes the way to estimate the number of supported prototypes in a specific cluster, and $\mu(\cdot)$ denotes the calculation of the number of query samples in $P_i$. Through regrouping the samples of each class into $n_{P_i^k}$ clusters, it yields the centroid $c^k=[c^k_1,...,c^k_{n_{P^k}}]$, where $n_{P^k}=\sum_{i=1}^{\left|P^k\right|}{n_{P_i^k}}$ signifies each split in $P^k$ that serves as the initialization points for the next iteration denoted as,
\begin{equation}
    P_i^k = K{\text -}means(\mathbf{z}, n_{P^k}, c^k).
\end{equation}

The above iteration will terminate when the conditions $\max{\delta(P_i)}=1$ and $\max{\mu(P)}<\alpha m$ are satisfied simultaneously, which also guarantees convergence under the previously mentioned assumption. Moreover, choosing an appropriate acceptance threshold for this clustering approach facilitates the optimal proficiency to discover novel categories. It is worth noting that when $\alpha\rightarrow+\infty$, this method will degenerate to continuously split the potential classes by $\delta(P_i) \ge 2$, which will undoubtedly lead to biases towards classes of existing support sets.

\subsection{Scalable clustering}
Compared to the UKC which is an evolution from standard K-means clustering, the proposed SHC exhibits relatively high computational complexity and memory usage due to its need to calculate and store distances between all pairs of data points, which grows exponentially with the size of the large-scale dataset, rendering it both computationally demanding and memory-intensive.

To mitigate the impact of the aforementioned issues, we employ a two-step strategy. In the first step, we randomly sample a smaller subset from the entire dataset, which is assumed to be representative since its distribution is consistent with that of the overall data distribution. The subsequent use of hierarchical clustering on this representative subset yields a set of potential clusters that can be viewed as initial prototypes. In the next step, these established prototypes serve as reference points for classifying the remaining data. Specifically, each data point will be merge to the tree which prototype is closest to it in the shared feature space. This strategy enables hierarchical clustering to be applied at scale, but its performance is influenced by the randomness of sampling and the quality of the initial clustering.

\section{Experiments}
\label{sec_exp}
\subsection{Experimental setup}

\begin{table}[!t]
\centering
\resizebox{0.49\textwidth}{!}{
\begin{tabular}{lcccccc}
\toprule
\multicolumn{2}{c}{} & \multicolumn{1}{c}{C100} & \multicolumn{1}{c}{I-100} & \multicolumn{1}{c}{CUB} & \multicolumn{1}{c}{Cars} & \multicolumn{1}{c}{Aircraft}\\ \cmidrule(l){1-7} 
\multirow{2}{*}{Training} & \multicolumn{1}{c}{$\vert \mathcal{Y}^{\text{Base}}\vert$} & \multicolumn{1}{c}{50} & \multicolumn{1}{c}{50} & \multicolumn{1}{c}{100} & \multicolumn{1}{c}{98} & \multicolumn{1}{c}{50}\\
& \multicolumn{1}{c}{$\vert\mathcal{D}^{\text{Base}}\vert$} & \multicolumn{1}{c}{25k} & \multicolumn{1}{c}{65k} & \multicolumn{1}{c}{3k} & \multicolumn{1}{c}{6.5k} & \multicolumn{1}{c}{4.4k}\\ \cmidrule(l){1-7} 
\multirow{2}{*}{Test} & \multicolumn{1}{c}{$\vert \mathcal{Y}^{\text{Novel}}\vert$} & \multicolumn{1}{c}{50} & \multicolumn{1}{c}{50} & \multicolumn{1}{c}{100} & \multicolumn{1}{c}{98} & \multicolumn{1}{c}{50}\\
& \multicolumn{1}{c}{$\vert\mathcal{D}^{\text{Novel}}\vert$} & \multicolumn{1}{c}{25k} & \multicolumn{1}{c}{65k} & \multicolumn{1}{c}{3k} & \multicolumn{1}{c}{8.1k} & \multicolumn{1}{c}{4.4k}\\
\bottomrule
\end{tabular}}
\vspace{-1ex}
\caption{Statistical comparison of data partitions (i.e., training and test) across C-100 (CIFAR-100), I-100 (ImageNet-100), CUB (CUB-200), Cars (StanfordCars) and Aircraft (FGVC-Aircraft).}
\label{tab_statistical}
\vspace{-3ex}
\end{table}

\noindent \textbf{Datasets.} We evaluate our methods on two well-known large-scale datasets: CIFAR-100 \cite{cifar} and ImageNet-100 \cite{imagenet}, as well as two fine-grained datasets, including CUB-200 \cite{cub} and Stanford Cars \cite{scar}. Each dataset is split into two sets: labeled data is used to train the model, and unlabeled data is used for testing. SSB (Semantic Shift Benchmark) \cite{vaze2022openset} provides a detailed evaluation dataset that includes precise ``semantic change axes'' and provides classifications for $\mathcal{D_U}$ and $\mathcal{D_L}$ in a semantically coherent manner. Statistics on the partitioning of adopted datasets are presented in Table \ref{tab_statistical}.

\noindent \textbf{Settings.} We mainly report results obtained with two different settings: 5-way 5-shot (5w5s) and 5-way 1-shot (5w1s). For better comparison, we set the number of new classes to 5 for both configurations (5n), denoted as $n_{new}$, with each class containing 15 images as the query set. Moreover, we initialize a real-time inference evaluation scenarios, which are in line with the objectives enabling the agent to freely switch between categorizing known and clustering novel classes. It is worth noting that for the real-time inference scenario, only one image from an individual class is allowed in the query set per episode.

A vision transformer \cite{ViT} (ViT-B-16) pre-trained on ImageNet \cite{imagenet} with DINO \cite{dino} is used for the feature extraction. Specifically, the outputs of $[CLS]$ token are treated as feature representation which is different from the use of a projection head that has the potential to introduce more biases. The initial learning rate is set to 0.01. 

\begin{table*}[!t]
\centering
\resizebox{\textwidth}{!}{
\begin{tabular}{ccccccccccccccccc}
\toprule
 &\multicolumn{1}{l}{} & \multicolumn{3}{c}{CIFAR-100} & \multicolumn{3}{c}{ImageNet-100} & \multicolumn{3}{c}{CUB-200} & \multicolumn{3}{c}{StanfordCars} & \multicolumn{3}{c}{FGVC-Aircraft} \\ \cmidrule(l){3-17} 
&\multicolumn{1}{l}{Methods} & All & Old & \multicolumn{1}{c|}{New} & All & Old & \multicolumn{1}{c|}{New} & All & Old & \multicolumn{1}{c|}{New} & All & Old & \multicolumn{1}{c|}{New} & All & Old & New \\ \cmidrule(l){1-17} 
\multirow{7}*{\rotatebox{90}{5way-5shot}}&\multicolumn{1}{l}{ProtoNet \cite{snell2017prototypical}} & 48.1 & \textbf{96.2} & \multicolumn{1}{c|}{-} & 47.7 & \textbf{95.4} & \multicolumn{1}{c|}{-} & 48.5 & \textbf{97.0} & \multicolumn{1}{c|}{-} & 42.9 & \textbf{85.8} & \multicolumn{1}{c|}{-} & 43.9 & \textbf{87.7} & - \\
&\multicolumn{1}{l}{RankStat \cite{autonovel1}} & 43.6 & 64.2 & \multicolumn{1}{c|}{23.0} & 41.1 & 59.1 & \multicolumn{1}{c|}{23.1} & 49.8 & 73.5 & \multicolumn{1}{c|}{26.1} & 39.4 & 57.8 & \multicolumn{1}{c|}{21.1} & 43.5 & 65.6 & 21.3 \\
&\multicolumn{1}{l}{SimGCD \cite{wen2023simgcd}} & 34.6 & 33.5 & \multicolumn{1}{c|}{35.7} & 31.1 & 30.9 & \multicolumn{1}{c|}{31.3} & 25.1 & 24.9 & \multicolumn{1}{c|}{25.4} & 17.1 & 16.2 & \multicolumn{1}{c|}{18.0} & 19.2 & 17.6 & 20.8 \\
&\multicolumn{1}{l}{OCD \cite{du2023on}} & 45.9 & 46.1 & \multicolumn{1}{c|}{45.6} & 13.5 & 0.1 & \multicolumn{1}{c|}{26.9} & 44.9 & 42.5 & \multicolumn{1}{c|}{47.2} & 37.3 & 34.5 & \multicolumn{1}{c|}{40.1} & 44.9 & 45.7 & \textbf{44.1} \\ 
&\multicolumn{1}{l}{GCD \cite{GCD}} & 63.8 & 92.7 & \multicolumn{1}{c|}{34.7} & 62.9 & 91.9 & \multicolumn{1}{c|}{33.9} & 68.7 & 95.0 & \multicolumn{1}{c|}{42.5} & 48.2 & 78.6 & \multicolumn{1}{c|}{17.8} & 50.0 & 80.5 & 19.5 \\\cmidrule(l){2-17} 
&\multicolumn{1}{l}{\textbf{FSNCD (SHC)}} & 73.2 & 90.7 & \multicolumn{1}{c|}{55.8} & 74.1 & 91.5 & \multicolumn{1}{c|}{56.7} & 73.6 & 94.5 & \multicolumn{1}{c|}{52.6} & \textbf{50.3} & 64.7 & \multicolumn{1}{c|}{35.9} & \textbf{51.0} & 71.2 & 30.8 \\
&\multicolumn{1}{l}{\textbf{FSNCD (UKC)}} & \textbf{84.3} & 90.9 & \multicolumn{1}{c|}{\textbf{77.8}} & \textbf{84.4} & 87.5 & \multicolumn{1}{c|}{\textbf{81.4}} & \textbf{85.8} & 92.5 & \multicolumn{1}{c|}{\textbf{79.1}} & 48.8 & 57.2 & \multicolumn{1}{c|}{\textbf{40.3}} & 49.5 & 57.3 & 41.8 \\ \cmidrule(l){1-17} 
\multirow{7}*{\rotatebox{90}{5way-1shot}} &\multicolumn{1}{l}{ProtoNet \cite{snell2017prototypical}} & 43.9 &\textbf{87.8} & \multicolumn{1}{c|}{-} & 42.6 & \textbf{85.2} & \multicolumn{1}{c|}{-} & 45.3 & \textbf{90.7} & \multicolumn{1}{c|}{-} & 33.0 & \textbf{66.0} & \multicolumn{1}{c|}{-} & 35.5 & \textbf{71.0} & - \\
&\multicolumn{1}{l}{RankStat \cite{autonovel1}} & 35.2 & 46.5 & \multicolumn{1}{c|}{23.8} & 33.8 & 43.9 & \multicolumn{1}{c|}{23.6} & 42.7 & 59.8 & \multicolumn{1}{c|}{25.7} & 29.3 & 37.6 & \multicolumn{1}{c|}{21.0} & 32.9 & 43.5 & 22.3 \\
&\multicolumn{1}{l}{SimGCD \cite{wen2023simgcd}} & 34.3 & 32.5 & \multicolumn{1}{c|}{36.1} & 31.2 & 29.5 & \multicolumn{1}{c|}{32.9} & 25.2 & 23.3 & \multicolumn{1}{c|}{27.1} & 17.3 & 15.7 & \multicolumn{1}{c|}{18.8} & 18.5 & 16.5 & 20.4 \\
&\multicolumn{1}{l}{OCD \cite{du2023on}} & 41.1 & 34.8 & \multicolumn{1}{c|}{47.5} & 12.3 & 0.1 & \multicolumn{1}{c|}{24.5} & 40.4 & 34.4 & \multicolumn{1}{c|}{46.5} & 34.3 & 28.8 & \multicolumn{1}{c|}{\textbf{39.8}} & 41.0 & 38.3 & \textbf{43.8} \\ 
&\multicolumn{1}{l}{GCD \cite{GCD}} & 63.4 & 77.3 & \multicolumn{1}{c|}{49.6} & 65.0 & 77.6 & \multicolumn{1}{c|}{52.5} & 66.7 & 83.1 & \multicolumn{1}{c|}{50.3} & 41.2 & 50.1 & \multicolumn{1}{c|}{32.3} & 44.2 & 55.1 & 33.3 \\ \cmidrule(l){2-17} 
&\multicolumn{1}{l}{\textbf{FSNCD (SHC)}} & 67.5 & 80.1 & \multicolumn{1}{c|}{54.8} & 69.9 & 82.1 & \multicolumn{1}{c|}{57.7} & 72.3 & 87.1 & \multicolumn{1}{c|}{57.5} & \textbf{42.4} & 48.2 & \multicolumn{1}{c|}{36.7} & \textbf{45.9} & 52.8 & 38.9 \\
&\multicolumn{1}{l}{\textbf{FSNCD (UKC)}} & \textbf{75.9} & 78.9 & \multicolumn{1}{c|}{\textbf{72.9}} & \textbf{76.6} & 76.5 & \multicolumn{1}{c|}{\textbf{76.6}} & \textbf{84.2} & 88.6 & \multicolumn{1}{c|}{\textbf{79.9}} & 38.9 & 41.5 & \multicolumn{1}{c|}{36.3} & 41.8 & 46.5 & 37.1 \\\bottomrule
\end{tabular}}
\vspace{-1.5ex}
 \caption{Main results of 15 query images for each class. The best result, is highlighted in \textbf{bold}. }
\vspace{-3ex}
\label{tab_main1}
\end{table*}

\noindent\textbf{Evaluation protocol.}
We assess the performance of the proposed new setting by measuring the accuracies of three folders, they are visible, new classes and overall precision. Concretely, it uses $\mathcal{\hat{Y}}_{All}$, $\mathcal{\hat{Y}}_{Old}$ and $\mathcal{\hat{Y}}_{New} = \mathcal{\hat{Y}}_{All} \setminus \mathcal{\hat{Y}}_{Old}$ to represent the predictions of examples from all, support set and new classes, respectively. Then, the metric for assessing new classes, commonly employed in the GCD approach, has been adapted to suit the FSNCD context. Formally,     
\begin{equation}
    ACC_{New} = \max_{p \in \mathcal{P}(\mathcal{\hat{Y}}{New})} \left( \frac{1}{M_{New}} \sum^{M_{New}}_{i=1} \mathbb{I}(y_i = p(\hat{y}_i)) \right),
\end{equation}
where $M_{New} = |\mathcal{\hat{Y}}_{New}|$, and $\mathcal{P}(\mathcal{\hat{Y}}_{New})$ defines how the predicted labels for test samples are matched to the true labels. This matching is achieved by using the Hungarian algorithm which identifies the permutation that minimizes the mismatch between predicted and true labels. 

Instead of directly comparing all predictions to the ground truth like GCD, we adopt a different approach. More specifically, we filter out samples from the same class as the support prototypes and then measure the matches between the predicted labels and prototypes by
\begin{equation}
    ACC_{Old}=\frac{1}{M}\sum^{M}_{i=1}{\mathbb{I}(y_i=p_{old}(\hat{y}_i))},
\end{equation}
where $p_{old}$ represents the one-to-one mapping from the label space of support prototypes $\mathcal{Y}_{Old}$ to $\mathcal{\hat{Y}}_{Old}$.


\noindent\textbf{FSNCD baselines.}
We present the following key benchmark methods as the strongest baselines for comparison with the proposed FSNCD. All methods employed ViT-B-16 pre-trained on ImageNet with DINO.

\textbf{(1) ProtoNet} \cite{snell2017prototypical} aims to classify samples into prototypes based on maximum cosine similarity, in contrast, our FSNCD identifies known classes while also considering how to discover novel classes.

\textbf{(2) OCD} \cite{du2023on} attempts to enhance the generalizability of GCD by using hash codes derived from disentangled features, while our FSNCD exploits a more general feature space to solve analogous generalization problems.

\textbf{(3) AutoNovel} \cite{autonovel1} uses feature ranking to determine whether a given sample belongs to positive pair, while we further extend its indexing ideas to the support prototypes. We use the two closest prototypes as category labels.

\textbf{(4) SimGCD} \cite{wen2023simgcd} attempts to utilize parametric classifier to solve GCD problem. Each episode of FSNCD can be regarded as a GCD problem with an extremely scarce sample size. We use the weights from Phase 1 and fine-tune using SimGCD to discover novel categories.

\textbf{(4) GCD} \cite{GCD} proposes to estimate the number of novel categories with accuracy on labeled samples, and discover categories with semi-supervised K-means. We treat each episode as a GCD task, performing estimation and clustering to discover novel categories.

\subsection{Main results}
In Table \ref{tab_main1}, we compare the performance of different methods on five distinct datasets. For simplicity, we denote the accuracy of samples in the support set as "old" and the accuracy of samples appearing in the query set but not in the support set as "new". Due to ProtoNet forcibly categorizing all samples into classes present in the support set, it lacks classification capability for new classes. We introduce ProtoNet into the comparison to observe the impact of new class discovery on the performance of old classes.

For methods based on K-means, GCD relies on K-means to estimate the number of categories and uses semi-supervised K-means clustering. However, it fails to discover new categories in coarse-grained datasets. In contrast, the method proposed in this paper does not require estimating the number of categories and achieves the best performance. Specifically, our method exhibits an average decrease of 9\% in old accuracy for SHC and 13\% for UKC compared to ProtoNet across multiple datasets. However, it demonstrates an average increase 5\% overall improvement across all classes for SHC and 10\% for UKC comparing to GCD.

OCD achieved nearly the same performance on various datasets and different task settings. Because the definition of categories is too strict, and the hash code of the same person usually implies multiple category information. As mentioned in OCD \cite{du2023on}, the length of the hash code is 12, so each sample can be divided into up to $2^{12}$ classes. With more support samples, we can obtain a more precise hash code for a certain category. The failure lies in the excessively strict category division and the lack of better utilization of the distribution of support features.

\begin{table*}[!t]
\centering
\resizebox{\textwidth}{!}{
\begin{tabular}{ccccccccccccccccc}
\toprule
 &\multicolumn{1}{l}{} & \multicolumn{3}{c}{CIFAR-100} & \multicolumn{3}{c}{ImageNet-100} & \multicolumn{3}{c}{CUB-200} & \multicolumn{3}{c}{StanfordCars} & \multicolumn{3}{c}{FGVC-Aircraft} \\ \cmidrule(l){3-17} 
&\multicolumn{1}{l}{Methods} & All & Old & \multicolumn{1}{c|}{New} & All & Old & \multicolumn{1}{c|}{New} & All & Old & \multicolumn{1}{c|}{New} & All & Old & \multicolumn{1}{c|}{New} & All & Old & New \\ \cmidrule(l){1-17} 
\multirow{5}*{\rotatebox{90}{Real-time}}&\multicolumn{1}{l}{ProtoNet \cite{snell2017prototypical}} & 48.1 & \textbf{96.2} & \multicolumn{1}{c|}{-} & 47.7 & \textbf{95.4} & \multicolumn{1}{c|}{-} & 48.5 & \textbf{97.0} & \multicolumn{1}{c|}{-} & 42.9 & \textbf{85.8} & \multicolumn{1}{c|}{-} & 43.9 & \textbf{87.7} & - \\
&\multicolumn{1}{l}{RankStat \cite{autonovel1}} & 59.7 & 63.6 & \multicolumn{1}{c|}{55.8} & 59.1 & 60.1 & \multicolumn{1}{c|}{58.0} & 61.9 & 73.7 & \multicolumn{1}{c|}{50.1} & 53.6 & 58.9 & \multicolumn{1}{c|}{48.3} & 54.4 & 64.3 & 44.5 \\
&\multicolumn{1}{l}{OCD \cite{du2023on}} & 66.2 & 45.2 & \multicolumn{1}{c|}{87.1} & 53.1 & 0.1 & \multicolumn{1}{c|}{\textbf{87.6}} & 67.3 & 41.5 & \multicolumn{1}{c|}{\textbf{93.0}} & 60.1 & 34.2 & \multicolumn{1}{c|}{86.0} & 59.2 & 33.2 & 85.2 \\ \cmidrule(l){2-17} 
&\multicolumn{1}{l}{\textbf{FSNCD (SHC)}} & 75.7 & 63.1 & \multicolumn{1}{c|}{\textbf{88.2}} & 82.3 & 73.0 & \multicolumn{1}{c|}{91.7} & 74.9 & 63.7 & \multicolumn{1}{c|}{86.1} & 57.1 & 17.7 & \multicolumn{1}{c|}{\textbf{96.4}} &56.0 & 15.7 & \textbf{96.2} \\
&\multicolumn{1}{l}{\textbf{FSNCD (UKC)}} & \textbf{78.3} & 70.0 & \multicolumn{1}{c|}{86.6} & \textbf{83.3} & 79.9 & \multicolumn{1}{c|}{86.7} & \textbf{78.1} & 74.8 & \multicolumn{1}{c|}{81.4} & \textbf{61.2} & 29.6 & \multicolumn{1}{c|}{92.8} & \textbf{61.0} & 30.4 & 91.6 \\
\bottomrule
\end{tabular}}
\vspace{-1.5ex}
\caption{Main result of real-time inference tasks. The best result, excluding ProtoNet, is highlighted in \textbf{bold}. }
\label{tab_real}
\vspace{-3ex}
\end{table*}

RankStat achieves optimal performance when encoding the top two ranks but limits the number of categories to 20, showing strong retention of classification capability for old classes as expected. Its strict partitioning of feature space into 20 regions based on support samples slightly reduces labeling effectiveness for new classes.

We further conducted tests for real-time inference. As shown in Table \ref{tab_real}, OCD easily identifies a sample as a new class, but it also tends to misclassify a substantial number of old classes as new. Although OCD achieves high accuracy in new class discovery, its utility is limited, indicating an excessive inclination toward categorizing instances as new classes. FSNCD with the SHC method achieves a more balanced performance in real-time inference, while FSNCD with the UKC exhibits lower capability in discovering new classes during real-time inference but still maintains excellent performance on old classes.

\begin{table}
\centering
\resizebox{0.48\textwidth}{!}{
\begin{tabular}{cccccccc}
\toprule
 &\multicolumn{1}{l}{} & \multicolumn{3}{c}{CIFAR-100} & \multicolumn{3}{c}{ImageNet-100} \\ \cmidrule(l){3-8} 
&\multicolumn{1}{l}{Methods} & All & Old & \multicolumn{1}{c|}{New} & All & Old & New \\ \cmidrule(l){1-8} 
\multirow{5}*{\rotatebox{90}{Large scale}} &\multicolumn{1}{l}{ProtoNet \cite{snell2017prototypical}} & 49.0 & \textbf{95.1} & \multicolumn{1}{c|}{-} & 47.7 & 95.4 & - \\
&\multicolumn{1}{l}{RankStat \cite{autonovel1}} & 57.8 & 61.3 & \multicolumn{1}{c|}{54.4} & 52.5 & 50.4 & 54.6 \\
&\multicolumn{1}{l}{OCD \cite{du2023on}} & 41.5 & 39.3 & \multicolumn{1}{c|}{43.7} & 25.2 & 22.7 & 27.6 \\ \cmidrule(l){2-8} 
&\multicolumn{1}{l}{\textbf{FSNCD (SHC)}} & 71.4 & 89.7 & \multicolumn{1}{c|}{53.1} & 76.4 & 94.6 & 58.2 \\
&\multicolumn{1}{l}{\textbf{FSNCD (UKC)}} & \textbf{89.3} & 92.2 & \multicolumn{1}{c|}{\textbf{86.3}} & \textbf{97.5} & \textbf{98.1} & \textbf{96.8}  \\ \bottomrule
\end{tabular}}
\vspace{-1.5ex}
\caption{Main result of large-scale dataset annotation tasks. The best result, excluding ProtoNet, is highlighted in \textbf{bold}.}
\label{tab_large}
\vspace{-2ex}
\end{table}

As shown in Table \ref{tab_large}, OCD struggles with large-scale dataset annotation due to its dispersed category distribution, resembling On-the-fly Category Discovery in accuracy. In contrast, UKC leverages sufficient samples to better capture data distributions, achieving superior performance across multiple metrics. Notably, discrepancies between Table \ref{tab_main1} and Table \ref{tab_large} are attributed to differing query sample sizes. Specifically, using centroids from small batches to classify large datasets enhances performance, aligning Cosine-trained models with Euclidean clustering. Additionally, in traditional GCD tasks, we found that the strategy of clustering and classification can significantly imporve performance. In extreme cases with limited query samples, it becomes challenging to determine whether a sample belongs to a new category, which is an inherent challenge in FSNCD.

\subsection{Ablation studies}

To validate the model's performance across various scenarios, we compared its performance under different settings. For more details on related parameter studies, please refer to the supplementary materials.

\noindent\textbf{Quantity of query.}
Due to the discovery of novel categories relying on clustering results, the model may be more sensitive to the distribution of data. Simultaneously, to ensure real-time reasoning capability, we investigated the impact of query quantity on model metrics. Initially, we set the task as 5w5s5n. 

\begin{figure}[!t]
    \centering
		\subfigure{\includegraphics[width=0.23\textwidth]{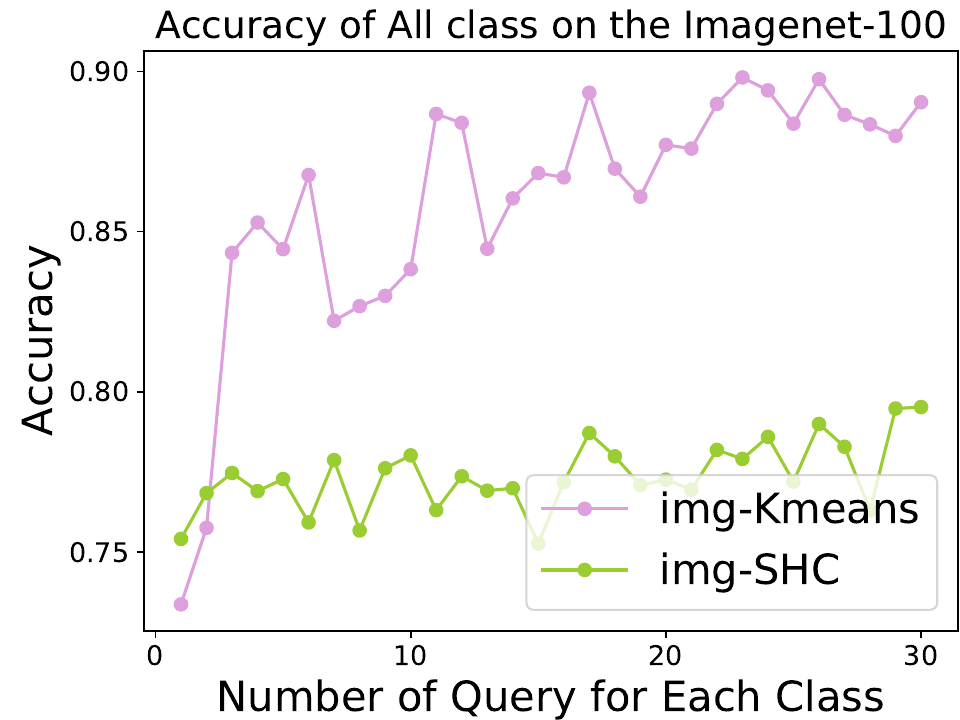}}
\subfigure{\includegraphics[width=0.23\textwidth]{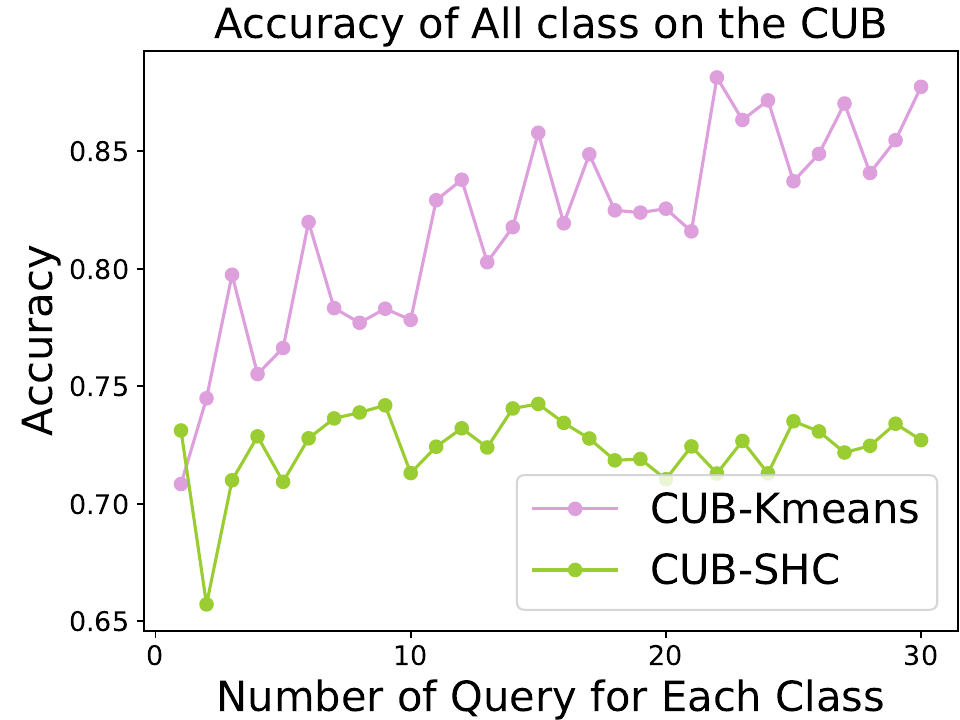}}%
\vspace{-2ex}
		\caption{Ablation studies for quantity of query using the 5w5s configuration. (a) The accuracy for Imagenet-100 on All classes. (b) The accuracy for CUB-200 All classes.}
    \label{abla_number}
    \vspace{-3ex}
\end{figure}

Fig. \ref{abla_number} illustrates the accuracy variation with the number of query samples in each class using the 5w5s configuration. The accuracy of UKC change curve indicates that as the number of query samples in each class increases, the data distribution tends to be more reasonable, leading to a continuous improvement. Conversely, due to the weak dependence of hierarchical clustering on data distribution, the accuracy of the SHC based method is more stable.

\noindent\textbf{Effects of different settings.} 
In response to distinct task configurations, we conducted further examinations on the outcomes associated with three specific task settings: (1) 5-way 5-shot 20-new, (2) 10-way 5-shot 20-new, and (3) 10-way 5-shot 40-new, and take 15 query images for each category. The results are shown in Table \ref{abla_task}. Given that UKC is contingent on the underlying data distribution, the results obtained from UKC exhibit a comparatively stable trend. Our approach demonstrates improvements across all three evaluation metrics as the number of supported sample categories increases. Notably, even with the introduction of additional novel classes, our method consistently maintains a commendable level of accuracy on pre-existing classes. This robust performance underscores the efficacy of our method in handling diverse task settings and accommodating the challenges posed by both an increased number of supported categories and the introduction of new ones.

\begin{table}[!t]
\centering
\resizebox{0.4\textwidth}{!}{
\begin{tabular}{cccccccc}
\toprule
 \multicolumn{2}{l}{} & \multicolumn{3}{c}{CUB-200} & \multicolumn{3}{c}{ImageNet-100} \\ \cmidrule(l){3-8}
\multicolumn{1}{l}{} & Method & All & Old & \multicolumn{1}{c|}{New} & All & Old & New \\ \cmidrule(l){1-8}
\multirow{2}{*}{(1)} & SHC & 52.4 & \textbf{91.9} & \multicolumn{1}{c|}{42.5} & 57.9 & \textbf{93.9} & 48.9 \\
& UKC & \textbf{61.2} & 86.5 & \multicolumn{1}{c|}{\textbf{54.9}} & \textbf{73.6} & 91.1 & \textbf{69.3} \\ \cmidrule{1-8}
\multirow{2}{*}{(2)} & SHC & 59.3 & \textbf{86.5} & \multicolumn{1}{c|}{45.7} & 72.1 & \textbf{88.7} & 63.8 \\
& UKC & \textbf{69.4} & 75.3 & \multicolumn{1}{c|}{\textbf{66.5}} & \textbf{83.4} & 87.0 & \textbf{81.6} \\ \cmidrule{1-8}
\multirow{2}{*}{(3)} & SHC & 48.8 & \textbf{84.1} & \multicolumn{1}{c|}{40.0} & 63.0 & \textbf{88.2} & 56.7\\
& UKC & \textbf{50.0} & 55.8 & \multicolumn{1}{c|}{\textbf{48.6}} & 79.4 & 80.4 & \textbf{79.1} \\
\bottomrule
\end{tabular}}
\vspace{-1.5ex}
\caption{Ablation study of different task. (1). 5-way 5-shot 20-new, (2). 10-way 5-shot 20-new, (3). 10-way 5-shot 40-new and both above tasks all use 15 query images of each category.}
\label{abla_task}
\vspace{-3ex}
\end{table}

\section{Conclusion}
\label{sec_con}
In this paper, we summarize the shortcomings of current Novel Category Discovery (NCD) tasks and Few-Shot Learning (FSL) tasks, considering FSL tasks that are closer to real-world scenarios, where new classes may appear in the query set. To address the aforementioned issues, we propose Few-shot Novel Category Discovery (FSNCD). In the inference stage, we introduce Semi-supervised Hierarchical Clustering (SHC) and Uncertainty-aware K-means Clustering (UKC) through supervised contrastive learning to represent learning in each episode and leverage them to classify old and cluster new classes accurately. We conduct comprehensive experiments on five datasets, validating the model's performance in general settings and its performance in extreme scenarios.

\bibliographystyle{named}
\bibliography{ijcai25}

\clearpage

\appendix

\section{Further Ablation studies}

At the meta-training phase, we only tune the last two blocks of the employed ViT model, and set the training epochs to 50. At the meta-test phase, we set the threshold $t$ in SHC to 2, meaning that a cluster is considered as the new class only when the number of classes exceeds 2. A threshold of 0 implies that it will be regarded as the new class even if a cluster has only one image. Although setting the threshold to 2 may seem meaningless, it is intended for our specific task setup to reduce the occurrence of individual outliers. In the ablation study, we explored the impact of the threshold, finding that a threshold of 0 could still achieve leading performance.

\noindent\textbf{Effects of $\alpha$.} In order to explore the impact of $\alpha$ on the results, we tested the influence of uncertainty $\alpha$ on UKC, where a higher $\alpha$ may lead to a higher acceptance level of the model for uneven feature distributions or category distributions. As shown on Table \ref{abla_alpha} that when $\alpha=1.4$, the performance of UKC on All class reaches its optimum. It can also be observed that with an increase in $\alpha$, prototypes from support samples mainly control category split, resulting in a gradual improvement in the classification performance on Old class, approaching the results of SHC.

\begin{table}[h]
\centering
\setlength{\tabcolsep}{5pt}
\begin{tabular}{ccccccc}
\toprule
 \multicolumn{1}{l}{} & \multicolumn{3}{c}{CUB-200} & \multicolumn{3}{c}{ImageNet-100} \\ \cmidrule(l){2-7} 
\multicolumn{1}{l}{$\alpha$} & All & Old & \multicolumn{1}{c|}{New} & All & Old & New \\ \cmidrule(l){1-7} 
\multicolumn{1}{l}{1.30} & 82.9 & 87.8 & \multicolumn{1}{c|}{78.0} & 87.7 & 90.1 & 82.3 \\
\multicolumn{1}{l}{1.35} & 83.5 & 89.2 & \multicolumn{1}{c|}{77.9} & 88.4 & 93.5 & \textbf{83.3} \\
\multicolumn{1}{l}{1.40} & \textbf{86.8} & 93.0 & \multicolumn{1}{c|}{\textbf{80.7}} & \textbf{88.9} & 94.8 & 83.1\\  
\multicolumn{1}{l}{1.45} & 77.9 & 92.1 & \multicolumn{1}{c|}{63.7} & 82.3 & 95.3 & 69.4 \\
\multicolumn{1}{l}{1.60} & 73.8 & \textbf{93.8} & \multicolumn{1}{c|}{53.8} & 74.1 & \textbf{96.2} & 51.8 \\
\bottomrule
\end{tabular}
\caption{Albation study for uncertainty $\alpha$ of UKC. The best result is highlighted in \textbf{bold}.}
\label{abla_alpha}
\end{table}

\noindent\textbf{Effects of threshold.} Semi-supervised Hierarchical Clustering often gives rise to the presence of numerous outliers. However, discerning whether these outliers represent instances from a newly introduced class characterized by limited sample size or merely outliers from pre-existing classes remains a challenge. Consequently, the introduction of a threshold becomes imperative to regulate the treatment of these outliers. In our exploration of the impact of the threshold on SHC clustering, we systematically adjusted the threshold values and conducted iterative testing on two distinct datasets. The comprehensive results are detailed in Table \ref{abla_threshold}. Notably, with the increment of the threshold, SHC exhibits a gradual improvement in its ability to accurately identify instances belonging to pre-existing classes. Strikingly, the highest accuracy for all instances as well as for new instances is achieved at the threshold value of $t=2$. This noteworthy observation indicates that SHC attains optimal sensitivity in managing outliers while simultaneously demonstrating heightened efficacy in handling instances from newly introduced classes at this specific threshold level.

\begin{table}[h]
\centering
\setlength{\tabcolsep}{7.5pt}
\begin{tabular}{ccccccc}
\toprule
 \multicolumn{1}{l}{} & \multicolumn{3}{c}{CUB-200} & \multicolumn{3}{c}{ImageNet-100} \\ \cmidrule(l){2-7} 
\multicolumn{1}{l}{t} & All & Old & \multicolumn{1}{c|}{New} & All & Old & New \\ \cmidrule(l){1-7} 
\multicolumn{1}{l}{0} & 69.1 & 87.7 & \multicolumn{1}{c|}{50.5} & 73.0 & 90.8 & 55.1 \\
\multicolumn{1}{l}{1} & 71.6 & 92.4 & \multicolumn{1}{c|}{51.2} & 76.3 & 95.6 & 57.2 \\
\multicolumn{1}{l}{2} & \textbf{73.5} & 93.8 & \multicolumn{1}{c|}{\textbf{54.3}} & \textbf{78.9} & 94.5 & \textbf{61.8}\\  
\multicolumn{1}{l}{3} & 73.2 & 94.6 & \multicolumn{1}{c|}{51.5} & 78.0 & 95.8 & 59.9 \\
\multicolumn{1}{l}{4} & 73.1 & \textbf{95.1} & \multicolumn{1}{c|}{50.8} & 78.7 & \textbf{96.1} & 59.2 \\
\bottomrule
\end{tabular}
\caption{Ablation study on the threshold of SHC. The best result is highlighted in \textbf{bold}.}
\label{abla_threshold}
\end{table}

\noindent\textbf{Effects of different model.} To demonstrate that the performance improvement of the proposed method does not stem from the effectiveness of the pre-training methods, we compared the impact of different models on performance. As shown in the Table \ref{abla_rn}, models without pre-training exhibit a significant decline in performance compared to those with pre-training, particularly the unpretrained ResNet-50. This is attributed to insufficient data and the excessive number of model parameters. Additionally, the lack of relevant information about new categories makes discovering them especially challenging, particularly because the semantic similarity between new and old categories in CIFAR-100 is relatively low, which may be a direction for future research.

\begin{table}[h]
\centering
\setlength{\tabcolsep}{5pt}
\resizebox{0.48\textwidth}{!}{
\begin{tabular}{ccccccccc}
\toprule
 \multicolumn{1}{l}{} & \multicolumn{1}{l}{} & \multicolumn{3}{c}{CUB-200} & \multicolumn{3}{c}{ImageNet-100} \\ \cmidrule(l){3-9} 
& Pretrain & Methods & All & Old & \multicolumn{1}{c|}{New} & All & Old & New \\ \cmidrule(l){1-9} 

\multirow{3}*{ViT-B-16} & \multicolumn{1}{c}{\multirow{3}*{\CheckmarkBold}} & ProtoNet & 48.1 & 96.2 & \multicolumn{1}{c|}{-} & 48.5 & 97.0 & - \\ 
&& SHC & 73.2 & 90.7 & \multicolumn{1}{c|}{55.8} & 73.6 & 94.5 & 52.6 \\
&& UKC & 84.3 & 90.9 & \multicolumn{1}{c|}{77.8} & 85.8 & 92.5 & 79.1 \\\midrule

\multirow{6}*{ResNet-18} & \multicolumn{1}{c}{\multirow{3}*{\XSolidBrush}} & ProtoNet & 35.4 & 70.8 & \multicolumn{1}{c|}{-} & 38.7 & 77.3 & - \\
&& SHC& 35.5 & 55.0 & \multicolumn{1}{c|}{15.9} & 44.5 & 62.8 & 26.2 \\
&& UKC& 45.7 & 56.5 & \multicolumn{1}{c|}{35.0} & 52.3 & 64.1 & 40.5 \\\cmidrule{2-9}

& \multicolumn{1}{c}{\multirow{3}*{\CheckmarkBold}} & ProtoNet & 46.2 & 92.5 & \multicolumn{1}{c|}{-} & 38.7 & 77.3 & - \\ 
&& SHC& 61.1 & 84.3 & \multicolumn{1}{c|}{37.8} & 69.4 & 91.1 & 47.7 \\
&& UKC& 69.6 & 88.3 & \multicolumn{1}{c|}{50.9} & 74.2 & 94.6 & 53.8 \\

\midrule
\multirow{6}*{ResNet-50} & \multicolumn{1}{c}{\multirow{3}*{\XSolidBrush}} & ProtoNet & 33.0 & 66.0 & \multicolumn{1}{c|}{-} & 27.1 & 54.2 & - \\
&& SHC& 36.5 & 64.2 & \multicolumn{1}{c|}{8.8} & 33.2 & 53.1 & 13.3 \\
&& UKC& 40.2 & 63.0 & \multicolumn{1}{c|}{17.4} & 34.5 & 52.5 & 16.5 \\\cmidrule{2-9}

& \multicolumn{1}{c}{\multirow{3}*{\CheckmarkBold}} & ProtoNet & 47.3 & 94.5 & \multicolumn{1}{c|}{-} & 47.9 & 95.8 & - \\ 
&& SHC& 52.7 & 92.8 & \multicolumn{1}{c|}{14.6} & 55.4 & 96.0 & 14.8 \\
&& UKC& 79.7 & 87.7 & \multicolumn{1}{c|}{71.6} & 81.4 & 88.2 & 74.7 \\

\bottomrule
\end{tabular}}
\caption{Albation study for different models.}
\label{abla_rn}
\end{table}

\begin{table}[h]
\centering
\resizebox{0.48\textwidth}{!}{
\begin{tabular}{ccccccc}
\toprule
\multirow{2}{*}{\makecell[l]{Transfer\\Direction}}& \multicolumn{3}{c}{SHC} & \multicolumn{3}{c}{UKC} \\ \cmidrule(l){2-7}
& All & Old & \multicolumn{1}{c|}{New} & All & Old & New \\ \cmidrule(l){1-7}
\multicolumn{1}{c||}{\hspace{0.08cm}$P_2\rightarrow P_1$} & 59.0 & 84.5 & \multicolumn{1}{c|}{33.4} & 67.9 & 72.5 & 63.2 \\
\multicolumn{1}{c||}{\hspace{0.18cm}$G\rightarrow P_1$} & 59.7 & 85.6 & \multicolumn{1}{c|}{33.8} & 70.8 & 80.5 & 61.0 \\ 
\multicolumn{1}{c||}{\hspace{0.08cm}$P_1\rightarrow P_1$} & 73.5 & 93.8 & \multicolumn{1}{c|}{54.3} & 86.8 & 93.0 & 80.7 \\\cmidrule(l){1-7}
\multicolumn{1}{c||}{\hspace{0.08cm}$P_1\rightarrow P_2$} & 29.0 & 57.8 & \multicolumn{1}{c|}{1.2} & 16.4 & 15.0 & 17.9 \\
\multicolumn{1}{c||}{\hspace{0.18cm}$G\rightarrow P_2$} & 50.3 & 64.6 & \multicolumn{1}{c|}{36.0} & 49.6 & 61.4 & 37.8 \\ 
\multicolumn{1}{c||}{\hspace{0.08cm}$P_2\rightarrow P_2$} & 52.0 & 63.3 & \multicolumn{1}{c|}{37.5} & 49.5 & 59.9 & 38.9 \\\cmidrule(l){1-7}
\multicolumn{1}{c||}{$P_1\rightarrow G$} & 77.7 & 96.7 & \multicolumn{1}{c|}{57.9} & 80.1 & 96.7 & 63.5 \\
\multicolumn{1}{c||}{$P_2\rightarrow G$} & 76.8 & 95.4 & \multicolumn{1}{c|}{58.3} & 88.7 & 94.6 & 82.8 \\ 
\multicolumn{1}{c||}{$\hspace{0.12cm}G\rightarrow G$} & 78.9 & 94.5 & \multicolumn{1}{c|}{61.8} & 88.9 & 94.8 & 83.1 \\
\bottomrule
\end{tabular}}
\caption{Effects of domain adaption. We explored different transfer tasks between different domains, where $P_1$: CUB-200, $P_2$: Stanford-Cars, and $G$: Imagenet-100.}
\label{alba_transfer}
\end{table}

\noindent\textbf{Effects of domain adaption.}
Following the conventions of common indicators in few-shot learning and drawing inspiration from the comparisons in the literature on Open Category Discovery (OCD) \cite{du2023on}, we introduced samples from different domains or data distributions during the testing phase to comprehensively assess the model's generalization performance. Specifically, we conducted transfer tasks on CUB-200, Stanford Cars, and ImageNet-100. CUB-200 (P1) and Stanford Cars (P2) represented two distinct specialized domains, while ImageNet-100 (G) represented a generic domain. We explored various transfer tasks between specialized and generic domains. Experimental results in Table \ref{alba_transfer} revealed a decline in performance for transfer tasks between specialized domains, indicating that despite the limited shared high-level semantics between Stanford Cars and CUB-200, the transfer from CUB-200 to Stanford Cars still yielded satisfactory results. Conversely, transfer tasks from CUB-200 to Stanford Cars exhibited poor results, suggesting that some linguistic features learned in Stanford Cars had good generalization performance on CUB-200.

It is noteworthy that for transfer tasks from specialized domains to a generic domain, as well as from a generic domain to specialized domains, no significant performance degradation was observed. On the contrary, the model continued to maintain robust performance in these transfer tasks. Particularly, in the transfer from specialized domains to a generic domain, the model's performance exhibited a trend of non-decrease.

\begin{figure*}[!t]
	\centering
	\includegraphics[width=0.99\textwidth]{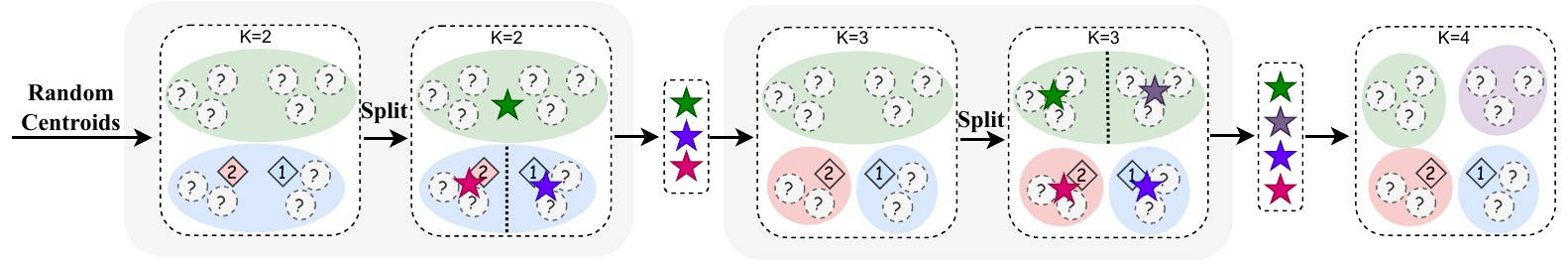} 
	\caption{An Example of Uncertainty-aware K-means Clustering (UKC) Based FSNCD. In the initialization stage, the model is clustered into $\left|\mathcal{Y_S}\right|$ classes with random centroids. The proposed criteria are employed to split clusters, extracting category centroids as heuristic information for the next iteration. This process continues until the specified conditions are met, resulting in the predictions of model.} 
	\label{fig_UKC}
\end{figure*}

\section{Visualization}
\begin{figure*}[h]
    \centering
    \subfigure{
        \includegraphics[width=0.23\textwidth]{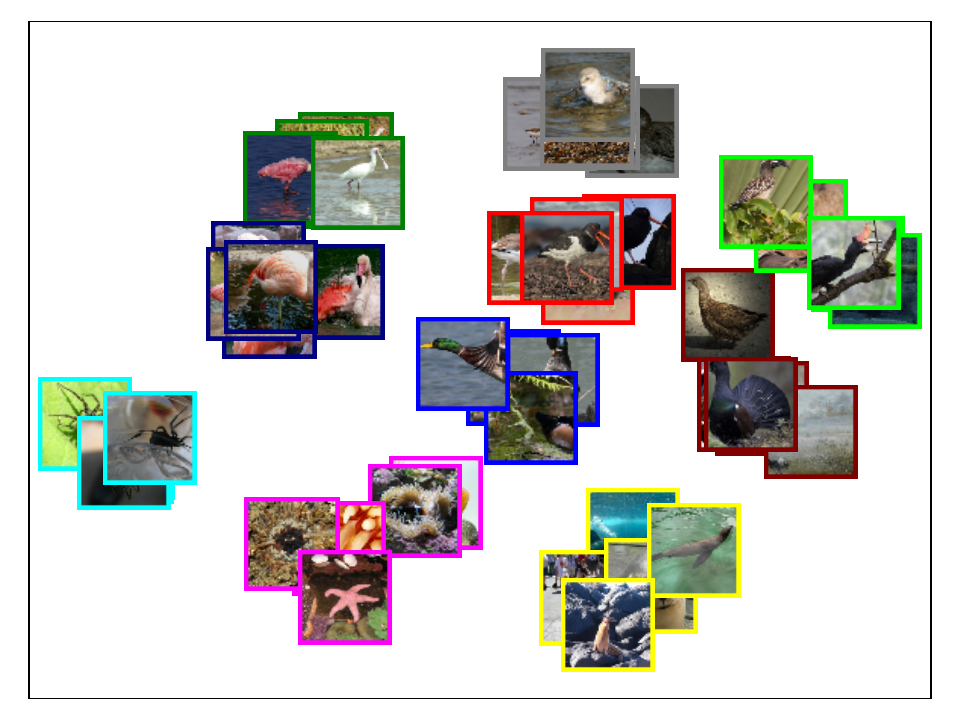}
    }
    \subfigure{
        \includegraphics[width=0.23\textwidth]{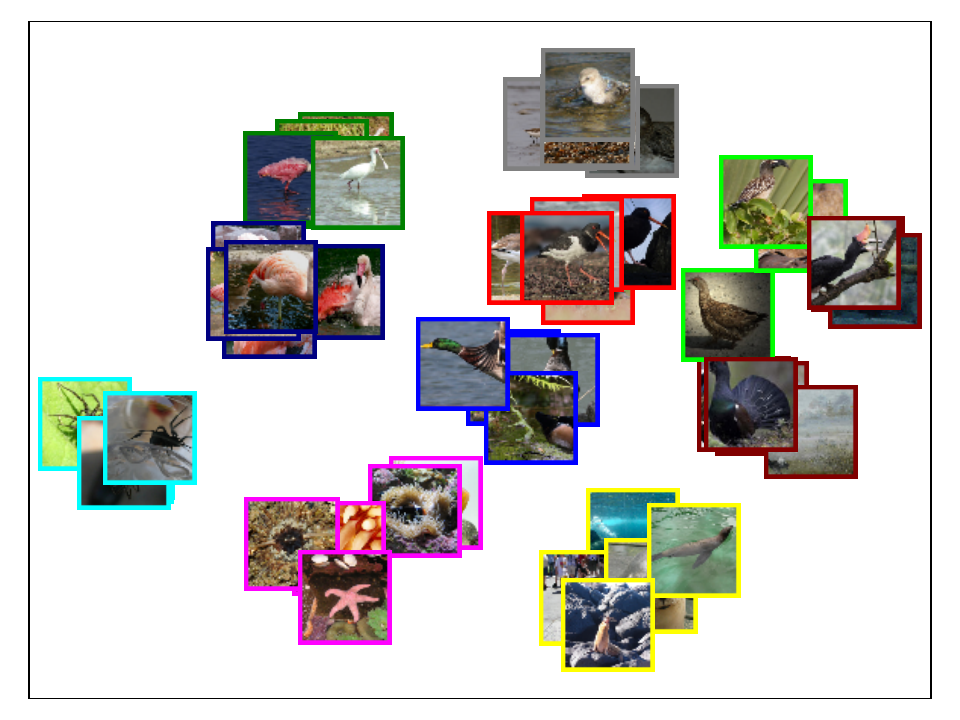}
    }
   \subfigure{
        \includegraphics[width=0.23\textwidth]{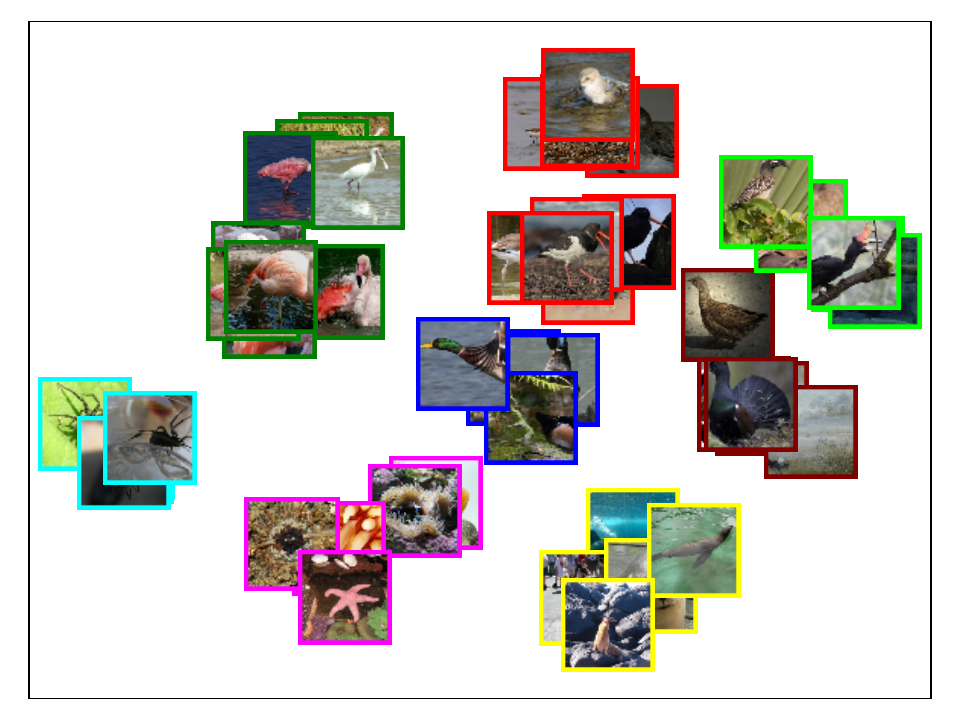}
    }
    \subfigure{
        \includegraphics[width=0.23\textwidth]{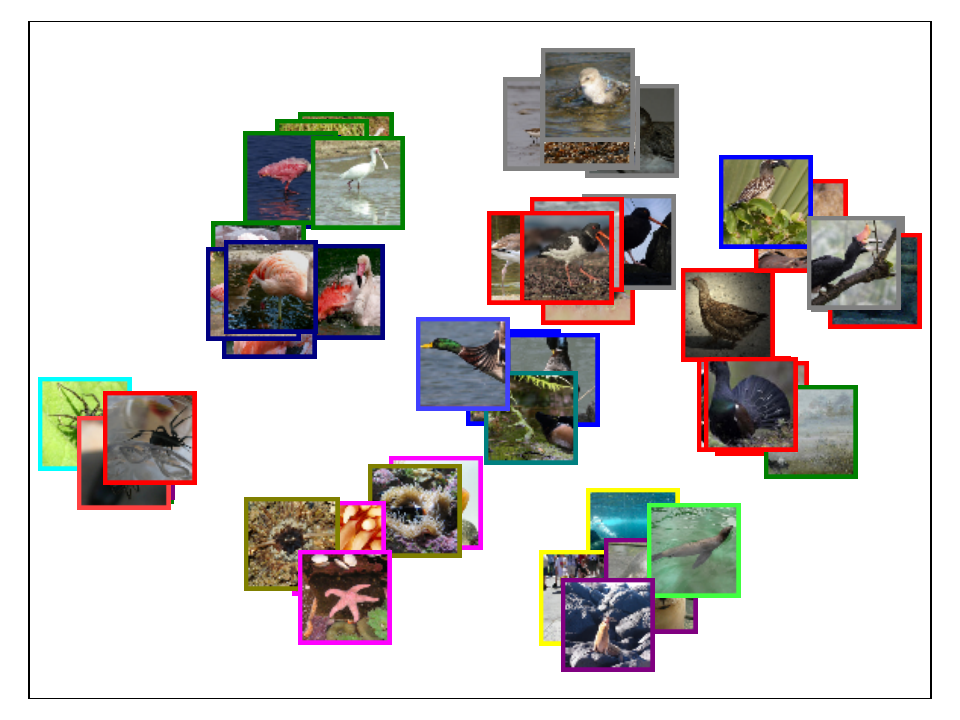}
    }

  \subfigure{
        \includegraphics[width=0.23\textwidth]{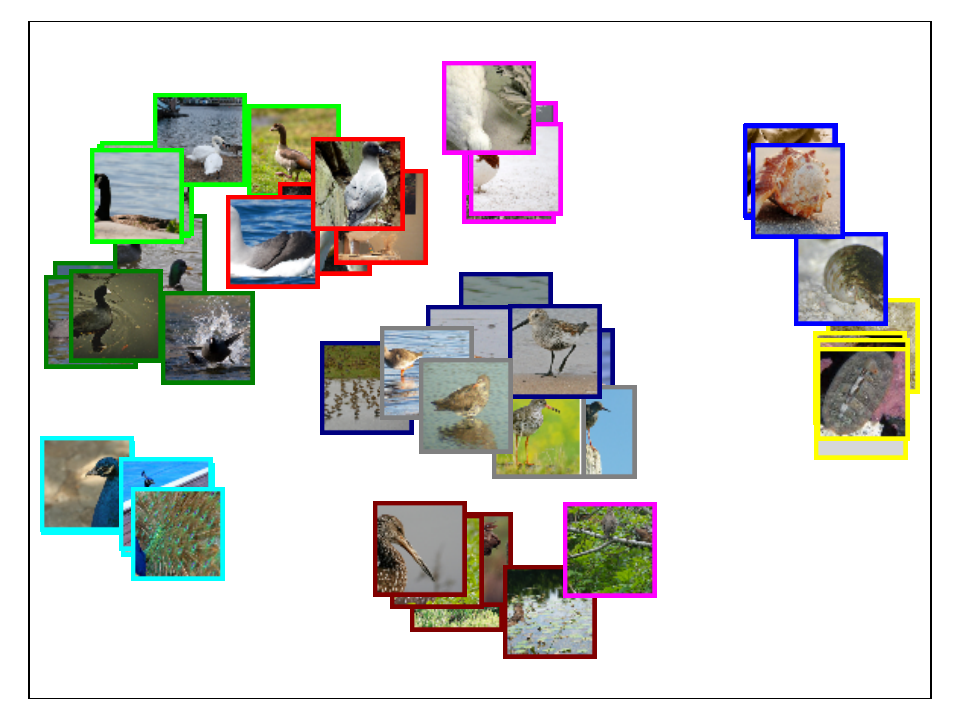}
    }
    \subfigure{
        \includegraphics[width=0.23\textwidth]{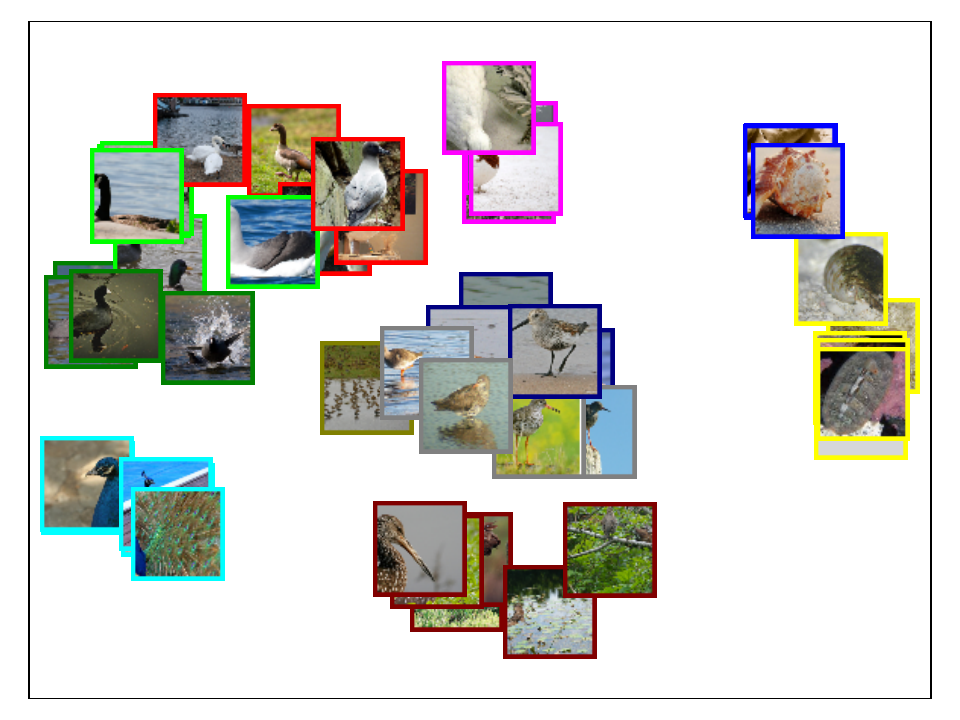}
   }
    \subfigure{
        \includegraphics[width=0.23\textwidth]{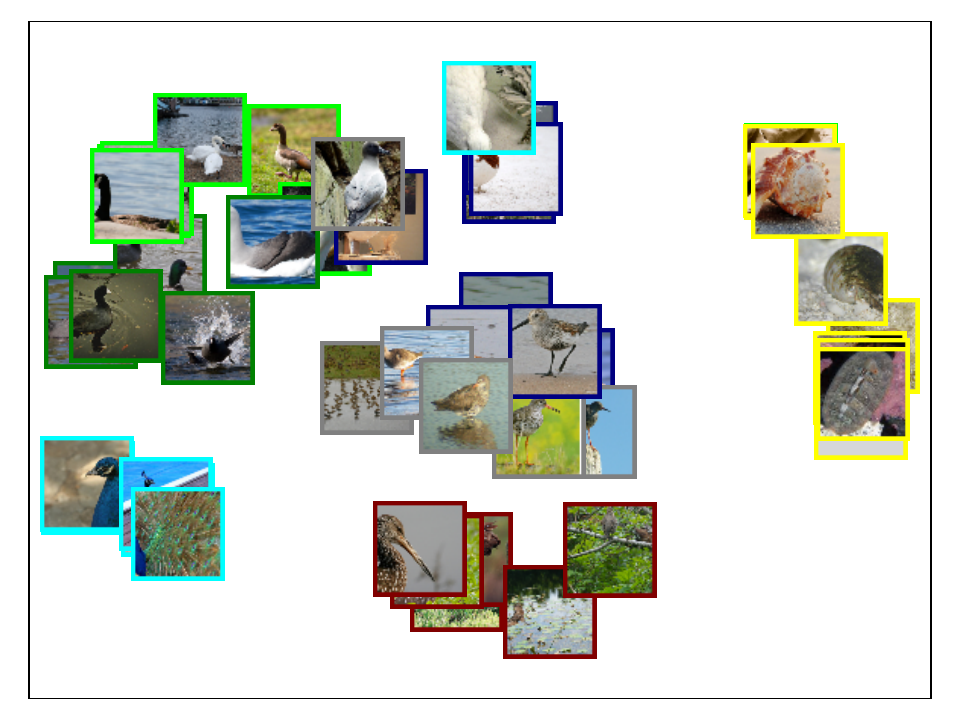}
    }
   \subfigure{
        \includegraphics[width=0.23\textwidth]{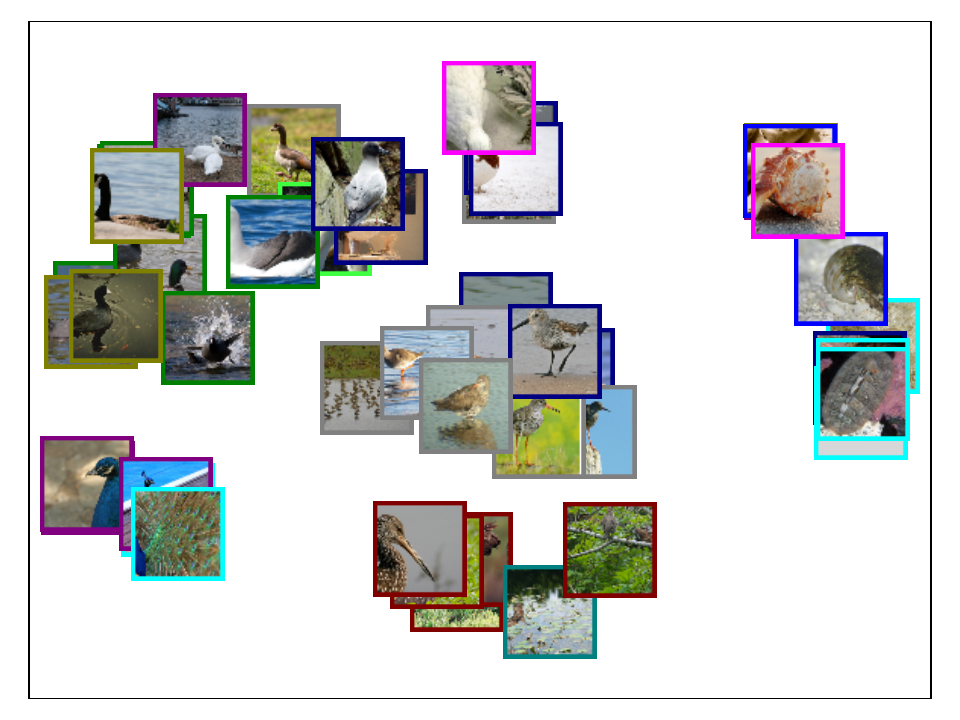}
   }

    \subfigure{
        \includegraphics[width=0.23\textwidth]{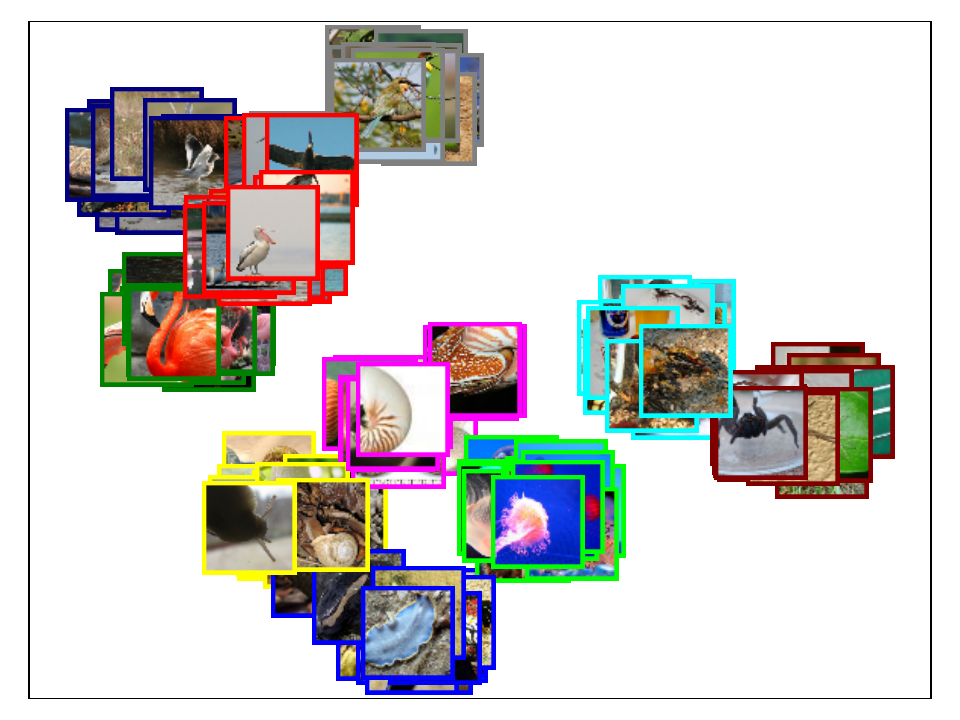}
    }
    \subfigure{
        \includegraphics[width=0.23\textwidth]{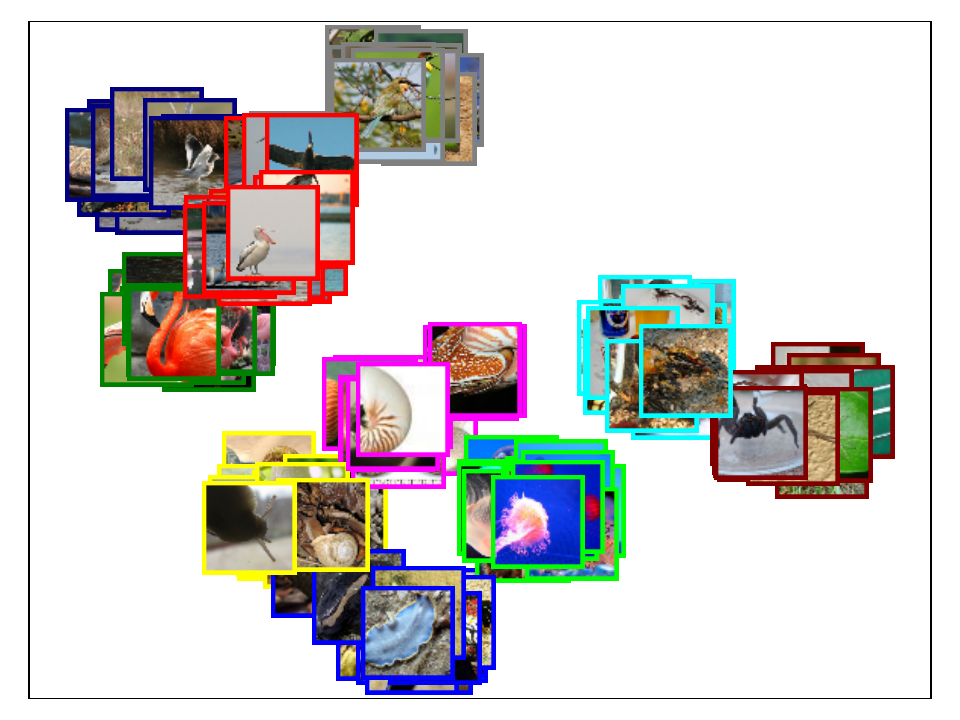}
    }
    \subfigure{
        \includegraphics[width=0.23\textwidth]{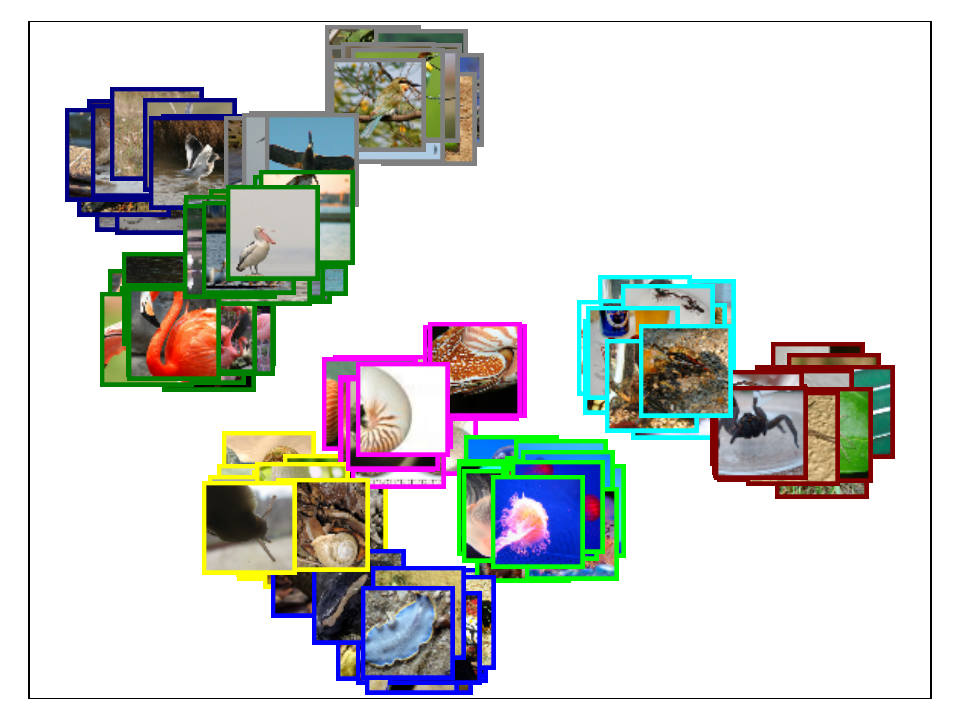}
    }
    \subfigure{
        \includegraphics[width=0.23\textwidth]{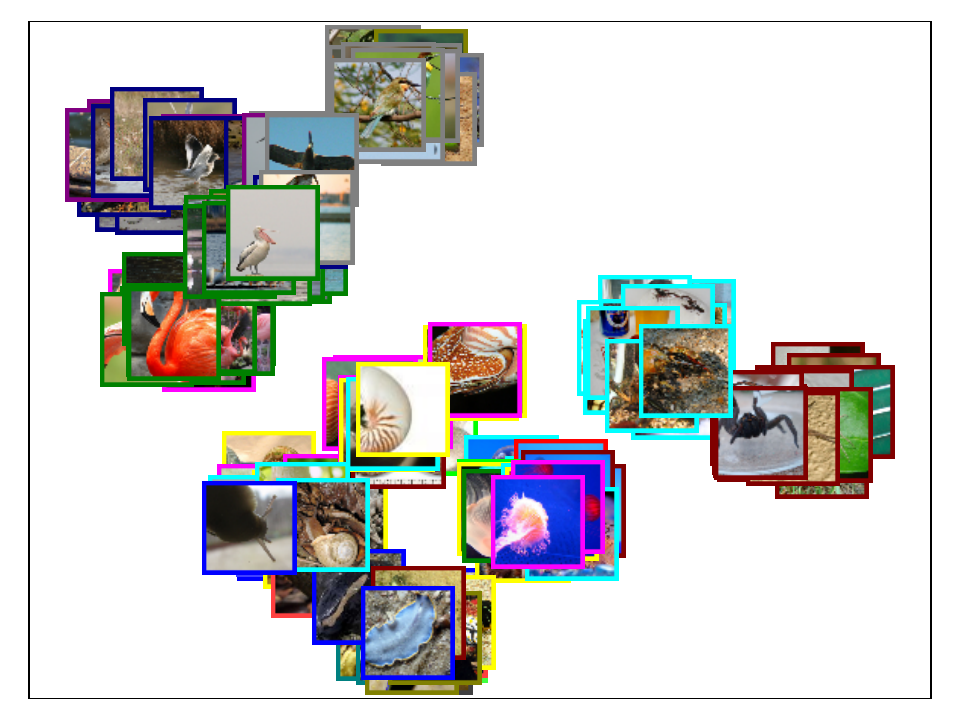}
    }
        
    \setcounter{subfigure}{0}
    \subfigure[Groundtruth]{
        \includegraphics[width=0.23\textwidth]{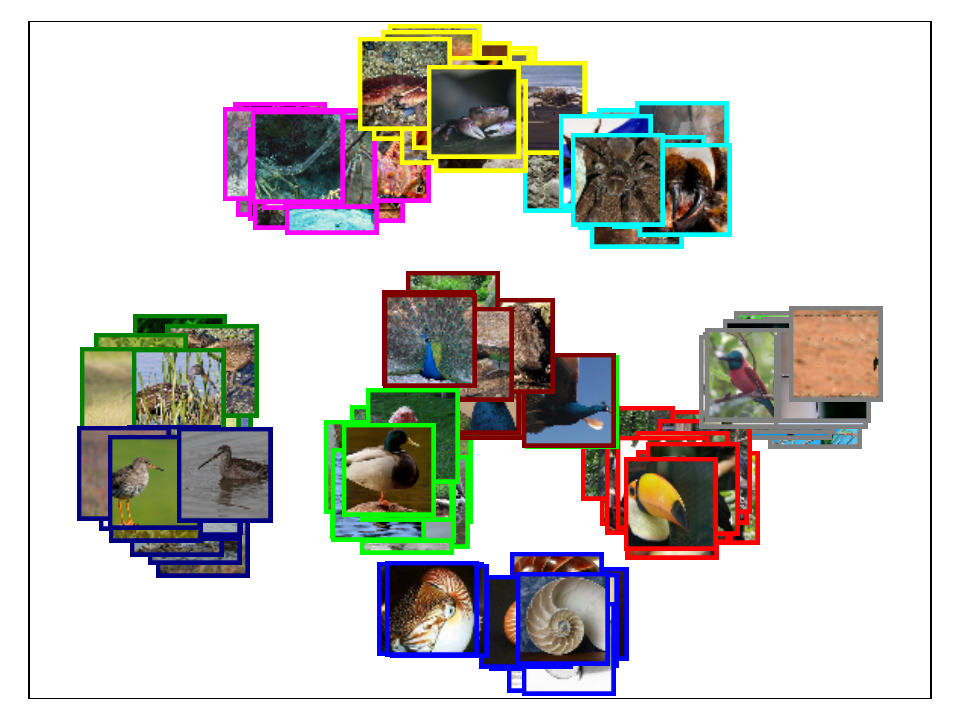}
    }
    \subfigure[UKC]{
        \includegraphics[width=0.23\textwidth]{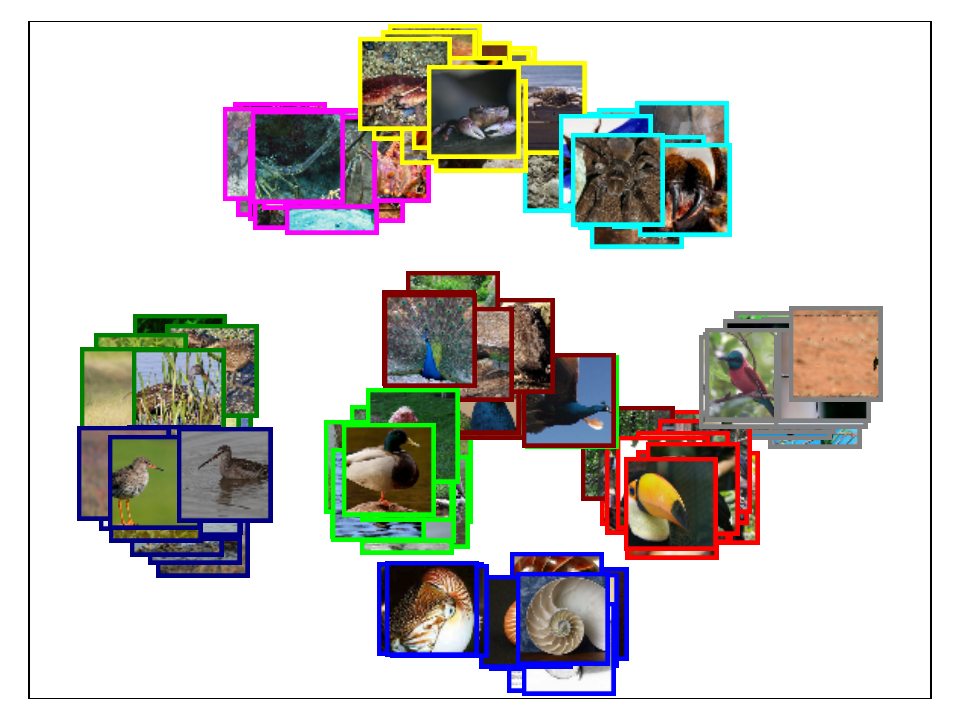}
    }
    \subfigure[SHC]{
        \includegraphics[width=0.23\textwidth]{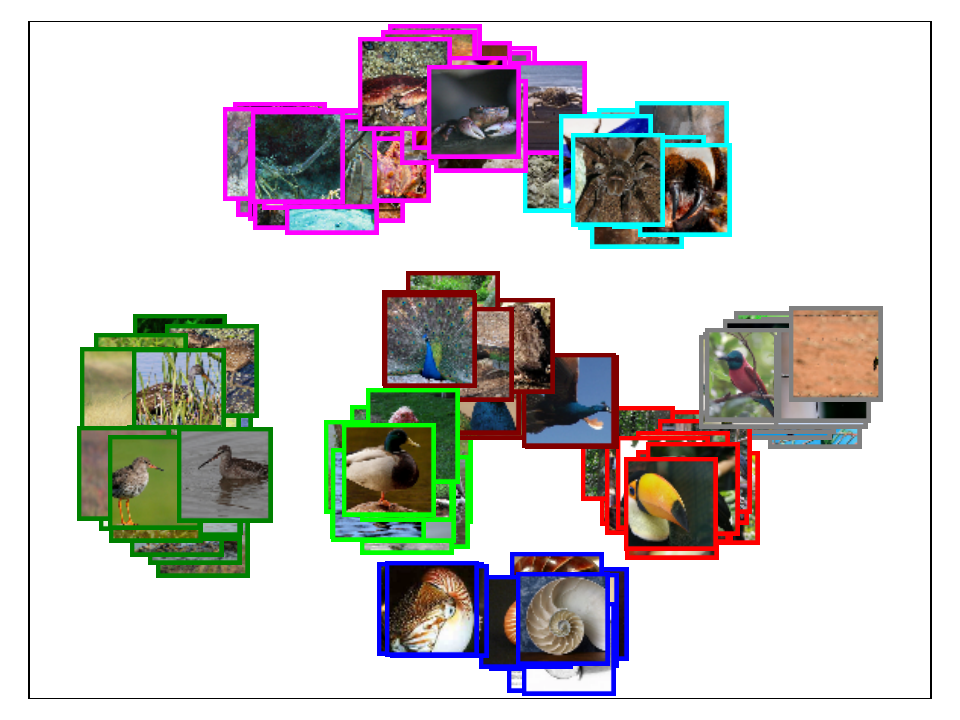}
   }
    \subfigure[Autonovel]{
        \includegraphics[width=0.23\textwidth]{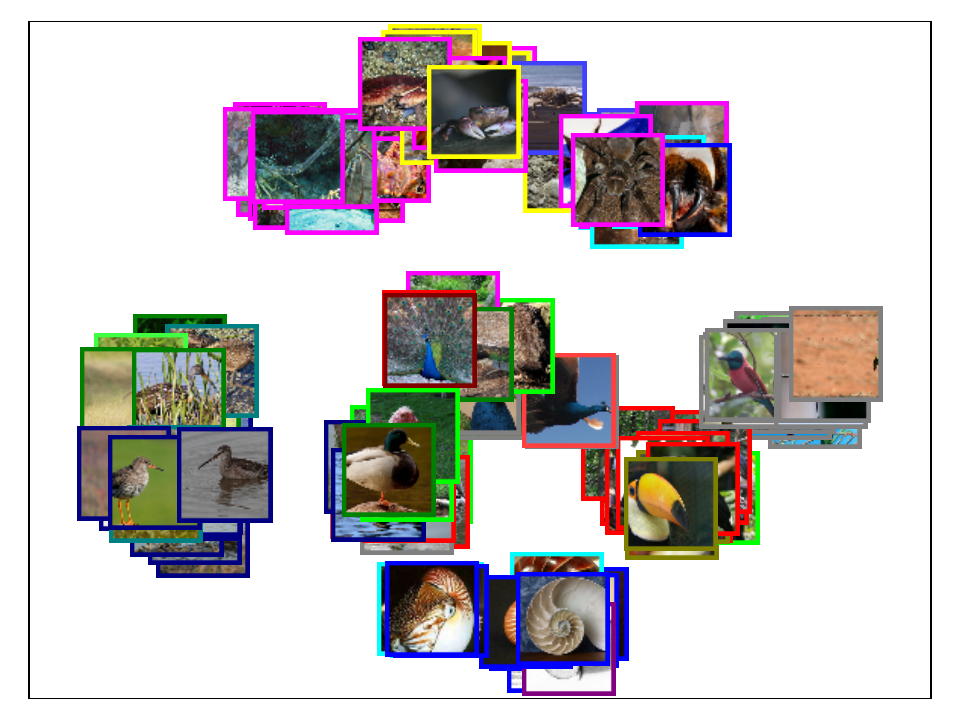}      
   }
    \caption{Visualization for different methods, with different colors indicating different categories.}
    \label{fid_vis}
\end{figure*}

To visualize the results of classification and clustering more intuitively, we visualized the classification performance of different methods on ImageNet with four different episode as shown in Fig. \ref{fid_vis}, where (a) represents the ground truth, (b) shows the results using UKC, (c) shows the results using SHC, and (d) shows the results using Autonovel \cite{autonovel1}. It can be observed that the classification results of UKC are the most accurate, followed by SHC. Due to the unreasonable partition of the feature space, Autonovel achieved relatively weak performance.

\begin{figure}[!t]
	\centering
	\includegraphics[width=0.49\textwidth]{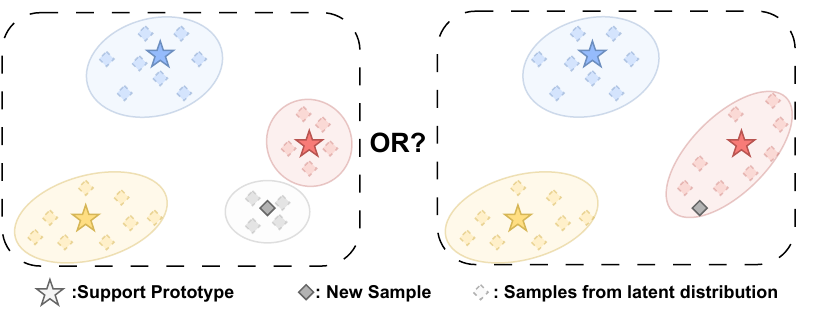} 
	\caption{Dilemmas faced by real-time reasoning tasks. The pentagon represents prototypes supporting samples, the solid line rhombus represents a query sample, and the dashed line rhombus represents a potential distribution that may exist.} 
	\label{fig_discussion}
\end{figure}

\section{Discussion}
\subsection{Further Explanation}

To provide a better explanation, we created the pseudo-code shown in Algo. \ref{algo_SHC} and a illustration for 2-way 1-shot 2-new task with UKC in Fig. \ref{fig_UKC}.

\begin{algorithm}[!t]
    \caption{Pseudo code of Semi-supervised Hierarchical Clustering (SHC) }
    \label{algo_SHC}
    \renewcommand{\algorithmicrequire}{\textbf{Input:}}
    \renewcommand{\algorithmicensure}{\textbf{Output:}}
    \begin{algorithmic}
        \REQUIRE{Support prototype $z_\mathcal{S}$ and Query feature $z_\mathcal{Q}$, cluster threshold $t$};
        \ENSURE{Potential cluster $P$};
        \STATE $C \gets z_\mathcal{S}\cup z_\mathcal{Q}$;
        \STATE $P, R = [\hspace{0.2cm}]$;
        \REPEAT
            \STATE $D \gets [d_{C_i, C_j}]_{\operatorname{len}(C)\times \operatorname{len}(C)}$;
            \STATE $i, j \gets \operatorname{argmax}_{i, j}D(i, j)$;
            \STATE $C_i \gets C_i \cup C_j$;
            \STATE $C \gets C $\textbackslash$ C_j$;
        \UNTIL{$C_i$ and $C_j$ both contains one of $z_\mathcal{S}$}.
        
        \FOR{$C_k \in C$}
            \IF {$\operatorname{len}(C_k)>t$}
                \STATE $P\gets C_k$;
            \ELSE
                \STATE $R\gets C_k$;
            \ENDIF
        \ENDFOR
        
        $D \gets [d_{P_i, R_j}]_{\operatorname{len}(P)\times \operatorname{len}(R)}$.
        
        \FOR{$i\in \operatorname{range}(\operatorname{len}(R))$}
            \STATE $j\gets \operatorname{argmax}_iD(:, i)$;
            \STATE $P(j)\gets R(i)$;
        \ENDFOR
    \end{algorithmic}
\end{algorithm}

Here is a simple example of FSNCD. If we give the children a few pictures of white foxes and a few pictures of squirrels before going to the zoo, they should know that they are foxes or squirrels when they meet them. However, it is difficult to inform children of all the animals, and when encountering flamingos, they should know that it is a new class, while flamingos always appear in flocks. And when there are enough of them, they should naturally find out that they belong to a class. Also based on the different granularity of the support set provided, there should be flexibility in determining whether the red fox belongs to the white fox.

From some perspectives, FSNCD is similar to out-of-distribution detection, and all we have are just a few support samples. How to make the most use of the distribution of support samples to achieve classification and discover new categories is crucial for FSNCD. As shown in Fig. \ref{fig_discussion}, in the scenario where there is only one query sample available and no additional query samples to assist in real-time inference tasks, it is difficult for us to determine whether a certain sample is a new class that is relatively close to the support prototype, or belongs to the same class but is far away from the support prototype.

The presence of two clustering algorithms provides insights from two perspectives to support the feasibility of the proposed FSNCD in open-set recognition scenarios where the underlying feature distribution of query examples is uncertain. The use of clustering alleviates the heavy reliance on the distribution of the query set by maximizing exploiting knowledge gained from the support set whose distribution is known in advance. 

\textbf{Trade-offs.}
To further clarify the characteristics of the two introduced clustering approaches, we will summarize and highlight their trade-offs in terms of \textit{stability} and \textit{inference speed}. As can be found from experimental results, UKC offers potentially higher performance but is sensitive to sample size and initialization, while SHC maintains more stable performance across varying conditions. 

\textbf{Opportunities.}
Offering both SHC and UKC options provides flexibility depending on whether performance or speed is prioritized for a specific task, but this comes at the cost of increased model complexity and potential maintenance overhead. How to effectively unify the framework will be the research we are committed to exploring. In addition, we found that SHC misclassified new classes as existing ones, leading to slightly underestimated class numbers. Our solution, in this situation, is that it will not be considered as a new class if the distance between the sample and prototype clusters is smaller than the shortest distance between prototypes in the high-level semantic space. We believe that the inflated performance could be improved if a metric like Greedy-Hungarian is used.

\end{document}